 \documentclass[twocolumn,twoside]{IEEEtran}
\usepackage{bm,amsmath,amssymb,amsfonts,graphicx,epsfig,amsthm,color}
\usepackage{graphicx,subcaption}

\usepackage[linesnumbered,ruled]{algorithm2e}

\newtheorem{proposition}{Proposition}
\newtheorem{lemma}{Lemma}

\newtheorem{theorem}{Theorem}
\newtheorem{definition}{Definition}
\newtheorem{assumption}{Assumption}
\theoremstyle{remark}\newtheorem{remark}{Remark}
\newcommand{\bld}[1]{\boldsymbol{#1}}
\usepackage{booktabs}
\usepackage{multirow}
\usepackage{dcolumn,array}
\allowdisplaybreaks

\begin{document}

\title{
 Learning while Respecting Privacy and Robustness to 
Distributional Uncertainties and Adversarial Data
}
\author{Alireza Sadeghi, Gang Wang, 
	 Meng Ma,
	 and Georgios B. Giannakis, \IEEEmembership{Fellow,~IEEE}
	\thanks{
		The work of A. Sadeghi, G. Wang, M. Ma, and G. B. Giannakis was supported in part by NSF grants 1711471, and 1901134. The authors are with the Digital Technology Center and the Department of Electrical and Computer Engineering, University of Minnesota, Minneapolis, MN 55455, USA (e-mail: sadeghi@umn.edu;  gangwang@umn.edu; maxxx971@umn.edu; georgios@umn.edu).}
}

\maketitle

\begin{abstract}
Data used to train machine learning models can be adversarial--maliciously constructed by adversaries to fool the model.~Challenge also arises by privacy, confidentiality, or due to legal constraints when data are geographically gathered and stored across multiple learners, some of which may hold even an ``anonymized'' or unreliable dataset. In this context, the distributionally robust optimization framework is considered for training a parametric model, both in centralized and federated learning settings.~The objective is to endow the trained model with robustness against adversarially manipulated input data, or, distributional uncertainties, such as mismatches between training and testing data distributions, or among datasets stored at different workers. To this aim, the data distribution is assumed unknown, and lies within a Wasserstein ball centered around the empirical data distribution.~This robust learning task entails an infinite-dimensional optimization problem, which is challenging.~Leveraging a strong duality result, a surrogate is obtained, for which three stochastic primal-dual algorithms are developed: i) stochastic proximal gradient descent with an $\epsilon$-accurate oracle, which invokes an oracle to solve the convex sub-problems; ii) stochastic proximal gradient descent-ascent, which approximates the solution of the convex sub-problems via a single gradient ascent step; and, iii) a distributionally robust federated learning algorithm, which solves the sub-problems locally at different workers where data are stored. Compared to the empirical risk minimization and federated learning methods, the proposed algorithms offer robustness with little computation overhead. Numerical tests using image datasets showcase the merits of the proposed algorithms under several existing adversarial attacks and distributional uncertainties.




\end{abstract}


\section{Introduction}
\label{sec:intro}
Recent advances in machine learning typically hinge on the premise that the training data are trustworthy, reliable, and representative of the testing data.~In practice however, data are usually generated and stored in geographically distributed devices (a.k.a., workers) each equipped with limited computing capability, due to data privacy, confidentiality, and/or cost associated with storing data in a centralized location \cite{federated_mag}.~Unfortunately,
the data quality at edge devices is not always guaranteed, where adversarially generated examples and distribution shift across users or between training and testing data distributions may be present \cite{fedlearn_challenges}.
Visually imperceptible perturbations to a dermatoscopic image of a benign mole can render the first-ever artificial intelligence (AI) diagnostic system approved by the U.S. Food and Drug Administration in 2018, to classify it as cancerous with 100\% confidence~\cite{mole}.~A stranger wearing pixelated sunglasses can fool even the most advanced facial recognition software in a home security system to mistake it for the homeowner~\cite{stranger}.~Hackers indeed manipulated readings of field devices and control centers of the Ukrainian supervisory control and data acquisition system to result in the first ever cyberattack-caused power outage in 2015 \cite{outage,tac2020wu,tsg2019wang}.~Examples of such failures in widely used AI-enabled safety- and security-critical systems today could put our lives and even national infrastructure at risk.





Recent research efforts have focused on devising defense strategies against adversarial attacks.~In general, these defense strategies can be categorized into two groups: attack detection, and attack recovery.~The former aims to identify whether a given input is adversarially perturbed \cite{gu2014arxiv, lu2017iccv}, while the latter trains a model with the goal of gaining robustness against such adversarial inputs \cite{guo2017countering, schmidt2018nips}, which is also the theme of this work.~To robustify learning models against adversarial data, a multitude of efforts have recently devised data pre-processing schemes
\cite{miyato2018regul,sheikholeslami2019minimum}, to signify data not adhering to postulated or nominal data generating models.~Adversarial training, on the other hand, injects such samples into the training dataset to gain robustness \cite{fellow2014adv}.~Generating adversarial samples was initially proposed in \cite{fellow2014adv}, by adding imperceptible well-crafted noise to clean input data.~Similar ideas have also been developed since then by e.g., \cite{madry2017towards,moosavi2017iccvpr, papernotattack}; see \cite{advsurvey1} for a recent survey.~In these contributions, certain optimization tasks are employed to craft adversarial perturbations.~Despite their empirical success, solving the optimization problems is challenging in general.~Furthermore, theoretical properties of these approaches have not been well understood, which limits the explainability of  obtained models.~In addition, one needs to judiciously tune hyper parameters of the attack model, which might be cumbersome in practice. 

On the other hand, data are typically generated and/or stored in multiple geographically distributed workers, where each worker may have a
different data distribution.~While keeping data localized due to e.g., data privacy, communication- and computation-overhead, the federated learning (FL) paradigm targets a global model by leveraging limited computational capabilities of the devices, which are coordinated by a central parameter server \cite{federated_mag}.~Existing works in FL have mainly focused on the communication and computation tradeoff by aggregating model updates from the learners; see e.g., \cite{fedavg,fedsanjabi,fedpoor,shlezinger2020uveqfed} and references therein.~Only a few works have explored robust FL.~Learning from non i.i.d data through e.g., sparsification \cite{fedrobustensemble} and
 ensemble of untrusted sources \cite{feduntrustedsources}, constitutes recent progress.~These methods are mostly heuristic, and have focused on appropriate aggregation schemes to gain robustness.~This context, motivates well a principled approach that {characterizes} the uncertainties associated with underlying data distributions.

%
%
%

%

 
 Tapping on a distributionally robust optimization perspective, this paper develops robust learning procedures that respects  robustness to distributional uncertainties and adversarial attacks.
The true data distribution is assumed unknown, but i.i.d. samples can be drawn.~Following \cite{sinha2017certify}, the adversarial input perturbations are modeled through a Wasserstein ball, and we are interested in a robust model that minimizes the worst-case expected loss over this Wasserstein ball of data distributions.~Unfortunately, the resulting task involves an infinite-dimensional optimization problem, which is challenging in general.
 Invoking a strong duality result, a tractable and equivalent unconstrained minimization problem
   is obtained, which requires solely the empirical data distribution.~To tackle this problem, this paper develops a stochastic proximal gradient descent (SPGD) algorithm with  $\epsilon$-accurate oracle,
    and its lightweight variant stochastic proximal gradient descent-ascent (SPGDA).~The first algorithm relies on the oracle to solve the emerging convex sub-problems to $\epsilon$-accuracy, while the second simply approximates its solution via a single gradient ascent step.~In addition, to accommodate (possibly untrusted) datasets distributed across multiple workers due to privacy issues and/or the communication overhead, a distributionally robust federated learning (DRFL) algorithm is further developed.
In a nutshell, the main contributions of this present paper are: 
\begin{itemize}
	\item A distributionally robust learning framework to endow machine learning models with robustness against adversarial input perturbations;
	
	
	\item Two scalable distributionally robust optimization algorithms with convergence guarantees; and,
	
	\item A distributionally robust federated learning implementation to account for untrusted and possibly anonymized data from multiple distributed sources. 
	

\end{itemize}

The rest of this paper is structured as follows. Problem formulation and its robust surrogate are provided in Section \ref{sec:probstate}. The proposed SPGD with $\epsilon$-accurate oracle and SPGDA algorithms with their convergence analyses are presented in Sections \ref{sec:spgdorac} and \ref{sec:spgda}, respectively. The DRFL implementation is discussed in Section \ref{sec:distrib}.~Numerical tests are given in Section~\ref{sec:experiments} with conclusions drawn in Section \ref{sec:concls}. Technical proofs are deferred to the Appendix.

\emph{Notation.} Bold lowercase letters denote
column vectors and calligraphic upper case letters are reserved for sets; $\mathbb{E}[\cdot]$ represents expectation; $\nabla$ denotes the gradient operator; $(\cdot)^\top$ denotes transposition, and $\|\bm x\|$ is the $2$-norm of the vector $\bm x$.

\section{Problem Statement}
\label{sec:probstate}
Consider the following standard regularized statistical learning problem
\begin{align}
& \underset{\bm{\theta}\in\Theta
}{{\rm min}}\;\;\mathbb{E}_{\bm{z}\sim P_0}\!\big[\ell(\bm{\theta};\bm{z})\big] + r(\bm{\theta})
\label{eq:reglearn}
\end{align}
where $\ell(\bm{\theta}; \bm{z})$ denotes the loss of a model parameterized by a set of unknowns $\bm{\theta}$ on a datum $\bm{z} =\! (\bm{x},y)$, with feature $\bm x$ and label $y$, drawn from some nominal distribution $\bm{z} \sim P_0$. Here, $\Theta$ denotes the feasible set for model parameters.   
To prevent over fitting or incorporate prior information, regularization term $r(\bm{\theta})$ is oftentimes added to the expected loss. Popular choices for the regularization term include $r(\bm{\theta}):=\beta \|\bm{\theta}\|_1^2$ or $\beta \|\bm{\theta}\|_2^2$, where $\beta \ge 0$ is a hyper-parameter controlling the importance of the regularization term.  

In practice, the nominal distribution $P_0$ is typically unknown. Instead, we are given some data samples $\{\bm{z}_n\}_{n=1}^N \! \sim \! \widehat P_0^{(N)}$ (a.k.a. training data), which are drawn i.i.d from $P_0$. Replacing $P_0$ with the so-called empirical distribution $\widehat P_0^{(N)}$ in \eqref{eq:reglearn}, we arrive at the empirical loss minimization 
\begin{align}
\underset{\bm{\theta}\in\Theta
}{{\rm min}}\;\; \bar{\mathbb{E}}_{\bm{z}\sim \widehat{P}^{(N)}_0}\!\big[\ell(\bm{\theta};\bm{z})\big] + r(\bm{\theta}) 
\label{eq:emlmin}
\end{align}
where $ \bar{\mathbb{E}}_{\bm{z}\sim \widehat{P}^{(N)}_0}\![\ell(\bm{\theta};\bm{z})] \!\!=\!\! N^{-1}  \sum_{n=1}^N \!\ell(\bm{\theta}; \bm{z}_n)$. 
Indeed, a variety of machine learning tasks can be formulated as \eqref{eq:emlmin}, including e.g., ridge- and Lasso regression, logistic regression, and even reinforcement learning problems. The resultant models obtained by solving \eqref{eq:emlmin} however, have been shown vulnerable to adversarially corrupted data in $\widehat{P}_0^{(N)}$. Further, the testing data distribution often deviates from the available $\widehat{P}_0^{(N)}$. Therefore, targeting an adversarially robust model against a set of distributions achievable by perturbing the underlying data distribution, the following formulation has been considered~\cite{sinha2017certify} 
\begin{equation}
\underset{\bm{\theta}\in\Theta
}{{\rm min}}\;\,\sup_{P\in\mathcal{P}}\,\mathbb{E}_{\bm{z}\sim P}\!\left[\ell(\bm{\theta};\bm{z})
\right] + r(\bm{\theta})\label{eq:minsup0}
\end{equation}
where $\mathcal P$ represents a set of distributions centered around the data generating distribution $\widehat{P}^{(N)}_0$. Compared with \eqref{eq:reglearn}, the worst-case formulation \eqref{eq:minsup0}, yields models ensuring reasonable performance across a continuum of distributions characterized by $\mathcal{P}$. In practice, different types of ambiguity sets $\mathcal{P}$ can be considered, and they lead to different robustness guarantees and computational complexities. Popular choices of $\mathcal P$ include momentum \cite{delage2010moment, wiesemann2014moment}, KL divergence \cite{hu2013kl}, statistical test \cite{bandi2014test}, and Wasserstein distance-based ambiguity sets \cite{bandi2014test,sinha2017certify}; see e.g., \cite{quantifying19blanchet} for a recent overview. Among all choices, it has been shown that Wasserstein ambiguity set $\mathcal{P}$ results in a tractable realization of \eqref{eq:minsup0}, in terms of strong duality result \cite{bandi2014test,sinha2017certify}, which also motivates this work. 


To formalize this, consider two probability measures $P$ and $Q$ supported on a set $\mathcal{Z}$, and let $\Pi(P,Q)$ be the set of all joint measures supported on $\mathcal{Z}^2$, with marginals $P$ and $Q$. Let $c: \mathcal{Z} \times \mathcal{Z}  \rightarrow [0, \infty)$ be a cost function measuring the cost of transporting a unit of mass from $\bm{z}$ in $P$ to another element $\bm{z}'$ in $Q$. The celebrated optimal transport
problem is given by \cite[page 111]{vinali08opttrans}
\begin{align}
W_c(P,Q) := \; \underset{\pi \in \Pi}{\inf} \, \mathbb{E}_\pi \big[ c(\bm{z},\bm{z}')\big]. 
\label{eq:wassdist}
\end{align}

\begin{remark}
If $c(\cdot,\cdot)$ satisfies the axioms of distance, then $W_c$ defines a distance on the space of probability measures. For instance, if $P$ and $Q$ are defined over a Polish space equipped with metric $d$, then choosing $c(\bm z, \bm z') = d^p(\bm z, \bm z')$ for some $p\in [1, \infty)$ asserts that $W_c^{1/p}(P,Q)$ is the well-known Wasserstein distance of order $p$ between probability measures $P$ and $Q$ \cite[Definition 6.1]{vinali08opttrans}.  
\end{remark}
For a given empirical distribution $\widehat{P}^{(N)}_0$, let us define uncertainty set $\mathcal{P}:= \{P| W_c(P,\widehat{P}^{(N)}_0) \le \rho\}$ to include all probability distributions having at most $\rho$-distance from $P_0^{(N)}$. Incorporating this ambiguity set into \eqref{eq:minsup0}, results in the following reformulation  
\begin{subequations}
\begin{equation}
\underset{\bm{\theta}\in\Theta
}{{\rm min}} \;\sup_{P}\,\mathbb{E}_{\bm{z}\sim P}\!\left[\ell(\bm{\theta};\bm{z})
\right] + r(\bm{\theta}) \label{eq:robforma}
\end{equation}
\begin{equation}
\qquad {\rm s.t.} \quad W_c(P,\widehat{P}^{(N)}_0) \le \rho. 
\label{eq:robformb}
\end{equation}
\label{eq:robform}
\end{subequations}

Observe that the inner supremum in \eqref{eq:robforma} runs over all joint probability measures $\pi$ on $\mathcal{Z}^2$ implicitly characterized by \eqref{eq:robformb}. Intuitively, directly solving this optimization over the infinite-dimensional space of distribution functions is challenging, if not impossible. Fortunately, for a broad range of losses as well as transportation cost functions, it has been shown that the inner maximization satisfies strong duality condition \cite{quantifying19blanchet}; that is the optimal objective of this inner maximization and its Lagrangian dual optimal objective are equal. In addition, the dual problem involves optimization over a one-dimensional dual variable. These two observations make it possible to solve \eqref{eq:minsup0} in the dual domain. To formally obtain a tractable surrogate to \eqref{eq:robform}, we make the following assumptions. 
\begin{assumption}
	\label{as:cfunc}
	The transportation cost function $c:\mathcal{Z}\times\mathcal{Z} \to [0, \infty)$, is a lower semi-continuous function satisfying $c(\bm{z}, \bm{z})=0$ for $\bm{z} \in \mathcal{Z}$\footnote{A simple example satisfying these constraints is the Euclidean distance $c(\bm z, \bm z') = \|\bm z - \bm z' \|$.}.
\end{assumption}

\begin{assumption}
	\label{as:lfunc}
	The loss function $\ell: \Theta \times \mathcal{Z} \to [0, \infty)$, is upper semi-continuous and integrable.
\end{assumption}

\noindent The following proposition provides a tractable surrogate for \eqref{eq:robform}, whose proof can be found in \cite[Theorem 1]{quantifying19blanchet}.  
\begin{proposition}
\label{prop:strongdual}
Let $\ell: \Theta \times \mathcal{Z} \rightarrow [0,\infty)$, and $c:\mathcal{Z} \times \mathcal{Z} \rightarrow [0, \infty)$ satisfy Assumptions $1$ and $2$, respectively. Then, for any given $\widehat{P}^{(N)}_0$, and $\rho > 0$, it holds that
\begin{align}
\sup_{P\in\mathcal{P}} & \,\mathbb{E}_{\bm{z}\sim P}\!\left[\ell(\bm{\theta};\bm{z})
\right] = \nonumber \\ 
& \inf_{\gamma \ge 0} \! \big\{  \bar{\mathbb{E}}_{\bm z \sim \widehat{P}^{(N)}_0} \!\big[\sup_{\bm \zeta \in {\mathcal Z}} \!\left\{ \ell(\bm \theta; \bm \zeta) \!-\! \gamma (c(\bm z, \bm \zeta)-\rho)\right\} \big] \big\}.
\label{eq:strongduality}
\end{align}
where $\mathcal{P}:= \left\{P| W_c(P,\widehat{P}^{(N)}_0) \le \rho \right\}$.
\end{proposition}

 

\begin{remark}
	Thanks to strong duality, the right-hand side in \eqref{eq:strongduality} simply is a univariate dual reformulation of the primal problem represented in the left-hand side. In sharp contrast with the primal formulation, the expectation in the dual domain is taken only over the empirical distribution $\widehat{P}^{(N)}_0$ rather than any $P \in \mathcal{P}$. In addition, since this reformulation circumvents the need for finding optimal coupling $\pi \in \Pi$ to form $\mathcal{P}$, and  characterizing the primal objective for all $P \in \mathcal{P}$, it is practically more convenient.
\end{remark}

Upon relying on Proposition \ref{prop:strongdual}, the following distributionally robust surrogate is obtained 
\begin{align}
\min_{\bm{\theta}\in \Theta} \inf_{\gamma \ge 0} \! \big\{  \bar{\mathbb{E}}_{\bm z \sim \widehat{P}^{(N)}_0} \!\big[\sup_{\bm \zeta \in {\mathcal Z}} \!\left\{ \ell(\bm \theta; \bm \zeta) \!+\! \gamma (\rho - c(\bm z, \bm \zeta))\right\} + r(\bm \theta) \big] \!\big\}.
\label{eq:robustdual}
\end{align}
   

\begin{remark}
\label{rm:minmax}
The robust surrogate in \eqref{eq:minsup0} falls into minimax (saddle-point) optimization. There are a vast majority of works to solve this kind of problems, see e.g., \cite{jordan2019minmax}. However,  the surrogate reformulation in \eqref{eq:robustdual}, requires that the supremum be solved separately for each sample $\bm{z}$, and the problem can not be handled through existing methods. 

A relaxed (hence suboptimal) version of \eqref{eq:robustdual} with a fixed $\gamma$ value 
has recently been studied in \cite{sinha2017certify}. Unfortunately, one has to select an appropriate $\gamma$ value using cross validation over a grid search that is also application dependent. Heuristically choosing a $\gamma$ does not guarantee optimality in solving the distributionally robust surrogate \eqref{eq:robustdual}. Clearly, the effect of heuristically selecting $\gamma$ is more pronounced when training deep neural networks. Instead, we advocate algorithms that optimize $\gamma$ and $\bm{\theta}$ simultaneously. 

Our approach to addressing this, relies on the structure of \eqref{eq:robustdual} to \textit{iteratively} update parameters $\bar{\bm \theta}:=[\bm \theta^\top~ \gamma]^\top$ and $\bm \zeta$. To end up with a differentiable function of $\bar{\bm \theta}$ after maximizing over $\bm \zeta$, Danskin's theorem requires the sup-problem to have a unique solution~\cite{bertsekas1997nonlinear}. For this reason, we design the inner maximization to involve a strongly concave objective function through the selection of a strongly convex transportation cost, such as $c(\bm{z}, \bm{z'}) :=\| \bm{z} - \bm{z'}\|^2_p$ for $p \ge 1$. For the maximization over $\bm \zeta$ to rely on a strongly concave objective, we let $\gamma \in \Gamma :=\{\gamma| \gamma >\gamma_0\}$, where $\gamma_0$ is large enough. Since $\gamma$ is the dual variable corresponding to the constraint in \eqref{eq:robform}, having $\gamma \in \Gamma$ is tantamount to tuning $\rho$ which in turn \emph{controls} the level of \emph{robustness}. Replacing $\gamma \ge 0$ in \eqref{eq:robustdual} with $\gamma \in \Gamma$, our \emph{robust learning model} is obtained as the solution of 
\begin{align}
\min_{\bm{\theta}\in \Theta}~\inf_{\gamma \in \Gamma}~\bar{\mathbb E}_{\bm{z} \sim \widehat{P}^{(T)}_0}\big[ \sup_{\bm{\zeta}\in\mathcal{Z}} \psi(\bm{\bar{\theta}}, \bm{\zeta}; \bm {z}) \big] + r(\bar{\bm{\theta}})
\label{eq:objective}
\end{align}
where $ \psi(\bm{\bar{\theta}}, \bm{\zeta}; \bm {z}):=\ell(\bm \theta;\bm \zeta) + \gamma (\rho -  c(\bm{z}, \bm \zeta)).$
Intuitively, input $\bm{z}$ in \eqref{eq:objective} is pre-processed by maximizing $\psi$ accounting for the adversarial perturbation. To iteratively solve our objective in \eqref{eq:objective}, ensuing sections provides efficient solvers under some mild conditions. Including, settings where we can solve every inner maximization (supremum) to $\epsilon$-optimality by an oracle.   
\end{remark}

Before developing our algorithms, we start by making several standard assumptions.

\begin{assumption}
\label{as:cstrongcvx}
$c(\bm z, \cdot)$ is $L_c$-Lipschitz and $\mu$-strongly convex function for any given $\bm{z} \in \mathcal{Z}$, with respect to the norm $\|\cdot\|$. 
\end{assumption}

\begin{assumption}
	\label{as:losslipschitz}
$\ell(\bm \theta; \bm z)$ satisfies the following Lipschitz smoothness conditions 
\begin{subequations}
	\label{eq:as2}
	\begin{align}
	& 
	\|\nabla_{\bm \theta} \ell(\bm{\theta}; \bm{z})- \nabla_{\bm{\theta}} \ell(\bm{\theta}';\bm{z})\|_{\ast} \le L_{\bm{\theta} \bm{\theta}} \|\bm{\theta}-\bm{\theta}'\|
	\label{eq:as2th}
	\\ 
	&
	\| \nabla_{\bm{\theta}} \ell(\bm{\theta; \bm{z}}) - \nabla_{\bm{\theta}} \ell(\bm{\theta}; \bm{z}')\|_\ast \le L_{\bm{\theta}\bm{z}} \|\bm{z}-\bm{z}'\| 
	\\ 
	&
	\label{eq:as2z}
	\| \nabla_{\bm{z}} \ell(\bm{\theta; \bm{z}}) - \nabla_{\bm{z}} \ell(\bm{\theta}; \bm{z}')\|_\ast \le L_{\bm{z}\bm{z}} \|\bm{z}-\bm{z}'\|
	\\
	&
	\label{eq:as2ztheta}
	\|\nabla_{\bm z} \ell(\bm{\theta}; \bm{z})- \nabla_{\bm{z}} \ell(\bm{\theta}';\bm{z})\|_{\ast} \le L_{\bm{z} \bm{\theta}} \|\bm{\theta}-\bm{\theta}'\|
	\end{align}
and it is continuously differentiable with respect to $\bm{\theta}$.
\end{subequations}
\end{assumption}

Assumption  \eqref{as:losslipschitz} would guarantee the supremum in \eqref{eq:robustdual} results in a smooth function of $\bar{\bm{\theta}}$, therefore one can execute gradient descent to update $\bm \theta$ upon solving the supremum. This will further help to provide convergence analysis of our proposed algorithms. To elaborate more on this, following lemma characterizes smoothness and gradient Lipschitz properties obtained upon solving the maximization problem in \eqref{eq:objective}. 

\begin{lemma}
\label{lem:smooth}
For each $\bm{z}\in\mathcal{Z}$, let us define 
$\bar{\psi} (\bar{\bm  \theta}; \bm{z})= \sup_{\bm{\zeta}} \psi (\bar{\bm  \theta}, \bm  \zeta; \bm{z})$ with $\bm{\zeta}_\ast(\bm{\bar{\theta}}; \bm{z}) := \arg \max_{\bm{\zeta} \in \mathcal{Z}} \psi(\bar{\bm  \theta}, \bm  \zeta; \bm{z})$. Then $\bar{\psi}(\cdot)$ is differentiable, and its gradient is $\nabla_{\bar{\bm{\theta}}} \bar{\psi}(\bar{\bm{\theta}};\bm{z}) = \nabla_{\bar{\bm{\theta}}} \psi(\bar{\bm{\theta}}, \bm{\zeta}_\ast(\bar{\bm{\theta}}; \bm{z});\bm{z})$. Moreover, the following conditions hold  
\begin{subequations}
\begin{equation}
\big \|\bm \zeta_\ast(\bar{\bm{\theta}}_1; \bm{z}) -  \bm \zeta_\ast(\bar{\bm{\theta}}_2; \bm{z}) \big \| \le \frac{L_{\bm{z}\bm{\theta}}}{\lambda}  \|\bm{\theta}_2 -\bm{\bm{\theta}}_1 \| + \frac{L_c}{\lambda}\, \| \gamma_2 -\gamma_1 \| \label{eq:lemmax}
\end{equation}
and
\begin{align}
\big\| \nabla_{\bar{\bm{\theta}}} \bar{\psi}(\bar{\bm{\theta}}_1; \bm{z}) -   \nabla_{\bar{\bm{\theta}}} \bar{\psi}(\bar{\bm{\theta}}_2; \bm{z})  \big\| \le \frac{ L_{\bm \theta z }L_c + L^2_c}{\lambda}\, & \| \gamma_2 -\gamma_1 \|  \nonumber \\ 
  +  (L_{\bm{\theta \theta}} + \frac{L_{\bm \theta z }L_{\bm{z}\bm{\theta}} + L_c L_{\bm{z}\bm{\theta}}}{\lambda})  \|\bm{\theta}_2 - & \, \bm{\bm{\theta}}_1 \|.  \label{eq:lemmapsi}
\end{align}
\end{subequations}
where $\gamma^{1,2}\in \Gamma$, and  $\psi(\bar{\bm{\theta}}, \cdot; \bm{z})$ is $\lambda$-strongly concave.
\end{lemma}
Proof: See Appendix \ref{app:lemm1} for the proof.

Lemma \ref{lem:smooth} paves the way for iteratively solving the surrogate optimization \eqref{eq:objective}, intuitively because it guarantees a differentiable and smooth objective upon solving the inner supremum to its optimum.
\begin{remark}
Equation \eqref{eq:lemmax} is of practical merits.~It ensures if $\bar{\bm{\theta}}^t\!=\![\bm{\theta}^t, \gamma^t]$ is updated with a small enough step size, the corresponding $\bm{\zeta}_\ast(\bm{\theta}^{t+1};\bm{z})$ is close enough to $\bm{\zeta}_\ast(\bm{\theta}^{t};\bm{z})$.~Building on this observation, instead of using an oracle to find the optimum $\bm{\zeta}_\ast(\bm{\theta}^{t+1};\bm{z})$, an $\epsilon$-accurate solution $\bm{\zeta}_\epsilon(\bm{\theta}^{t+1};\bm{z})$ suffices to obtain similar performance.~Enticingly, this circumvents the need to find the optimum for the inner maximization per iteration, which could be computationally demanding. 
\end{remark}




\section{Stochastic Proximal Gradient Descent with $\epsilon$-accurate Oracle}
\label{sec:spgdorac}

\begin{algorithm}[t]
	\SetKwInOut{Input}{Input}
	\SetKwInOut{Output}{Output}
	\SetKw{Set}{Set}	
	\Input{Initial guess $\bm{\bar{\theta}}^0$, step size sequence $\{\alpha_t >0 \}_{t=0}^{T}$, $\epsilon$-accurate oracle}
	
		\For{$t = 1, \ldots, T$}
		{ {Draw i.i.d samples $\{\bm{z}_n\}_{n=1}^N$} \\ 
			{Find $\epsilon$-optimizer $\bm{\zeta}_\epsilon(\bar{\bm{\theta}^t}; \bm{z}_n)$ via the oracle} \\
			{Update:  \newline $\bar{\bm{\theta}}^{t+1} = {\rm prox}_{\alpha_t r} \Big[ \bar{\bm{\theta}}^t-\frac{\alpha_t}{N}\sum_{n=1}^N \nabla_{\bar{\bm{\theta}}} \psi (\bar{\bm{\theta}}, {\bm \zeta}_\epsilon(\bar{\bm{\theta}}^t; \bm{z}_n); \bm{z}_n)\big|_{\bar{\bm{\theta}}=\bar{\bm{\theta}}^t} \Big]$}
		}	  
	\caption{SPGD with $\epsilon$-accurate oracle}
	\label{alg:spgd}
\end{algorithm}
A standard approach to solving regularized optimization problems
is the proximal gradient algorithm.~In this section,
we develop a variant of this method to tackle the robust surrogate \eqref{eq:objective}. For convenience, let us define
\begin{equation}
f(\bm{\theta}, \gamma) := 
{\mathbb{E}} \;
 \!\big[\sup_{\bm{\zeta} \in {\mathcal Z}} \!\left\{ \ell(\bm \theta; \bm{\zeta}) \!+\! \gamma (\rho - c(\bm z, \bm{\zeta}) ) \right\} \big] \label{eq:fdef}
\end{equation}
and rewrite our objective as follows
\begin{align}
\min_{\bm{\theta}\in \Theta} \inf_{\gamma \in \Gamma} \; F(\bm{\theta}, \gamma) := f(\bm{\theta}, \gamma) + r(\bm{\theta}) \label{eq:sorrogate}
\end{align}
where $f(\bm{\theta}, \gamma)$ is a smooth function defined in \eqref{eq:fdef}, and $r(\bm{\cdot})$ is a non-smooth and convex regularizer, such as $\ell_1$-norm. 
With slight abuse of notation, upon introducing $\bar{\bm \theta}:=[\bm \theta~\gamma]$,  we define  $f(\bar{\bm \theta}):= f(\bm \theta, \gamma)$ and $F(\bar{\bm \theta}):= F(\bm \theta, \gamma)$. The proximal gradient algorithm updates $\bar{\bm{\theta}}^t$, as follows 
\begin{equation*}
\bar{\bm \theta}^{t+1} = \arg \min_{\bm \theta} \, \alpha_t r(\bm \theta) + \alpha_t \big\langle \bm \theta - \bar{\bm \theta}^t, \bm g (\bar{\bm \theta}^t) \big \rangle + \frac{1}{2} \big\|\bm \theta - \bar{\bm \theta}^{t}\big\|^2 
\end{equation*}
where $\bm g (\bar{\bm \theta}^t) := \nabla f(\bar{\bm{\theta}})|_{\bar{\bm{\theta}} = \bar{\bm{\theta}^t}}$, and $ \alpha_t > 0$ is some step size. Usually, this update is expressed in a compact form 
\begin{equation}
\bar{\bm \theta}^{t+1} = \textrm{prox}_{\alpha_t r} \big[\bar{\bm \theta}^{t} - \alpha_t \bm g(\bar{\bm  \theta}^{t})\big]\label{eq:thetaupdt}
\end{equation}
where the proximal gradient operator is defined as 
\begin{equation}
\textrm{prox}_{\alpha r} [\bm  v] := \arg  \min_{\bm  \theta} \; \alpha r(\bm  \theta) + \frac{1}{2}\|\bm  \theta - \bm  v\|^2. \label{eq:proximal}
\end{equation}
The working assumption is that this optimization problem can be solved efficiently using off-the-shelf solvers.

Starting from some guess $\bar{\bm \theta}^0$, the proposed SPGD with $\epsilon$-accurate oracle executes two steps at every iteration $t=1,$ $2,\ldots$. First, it relies on an $\epsilon$-accurate maximum oracle to solve the inner problem $\sup_{\bm \zeta \in {\mathcal Z}} \! \{ \ell(\bm \theta^t; \bm \zeta) \!-\! \gamma^t c(\bm z, \bm \zeta)\}$ for randomly drawn samples $\{\bm{z}_n\}_{n=1}^{N}$ to yield $\epsilon$-optimal  $\bm{\zeta}_\epsilon(\bar{\bm{\theta}}^t,\bm{z}_n)$ with the corresponding objective values $\psi(\bar{\bm{\theta}}^t,\bm  \zeta_{\epsilon}(\bar{\bm \theta}^t, \bm  z_n); \bm  z_n)$. Next, $\bar{\bm{\theta}}^t$ is updated using a stochastic proximal gradient step as follows
\begin{equation}
\bar{\bm{\theta}}^{t+1} = {\rm prox}_{\alpha_t r} \Big[ \bar{\bm{\theta}}^t-\frac{\alpha_t}{N}\sum_{n=1}^N \nabla_{\bar{\bm{\theta}}} \psi (\bar{\bm{\theta}}, \bm \zeta_\epsilon(\bar{\bm{\theta}}^t; \bm{z}_n); \bm{z}_n) \Big]. \nonumber
\end{equation}

For implementation, the proposed SPGD algorithm with $\epsilon$-accurate oracle is summarized in Alg. \ref{alg:spgd}.~Convergence performance of this algorithm is analyzed in the ensuing subsection. 


\subsection{Convergence of SPGD with $\epsilon$-accurate oracle} 
\label{suc:convspgd}
In general, the postulated model is nonlinear, and the robust surrogate objective function $F(\bar{\bm{\theta}})$ is nonconvex.~In this section, we characterize the convergence performance of Alg. \ref{alg:spgd} to a stationary point.~However, due to the nonconvexity and nonsmoothness, stationary points are defined in the sense of Fr\`echet subgradient.~Specifically, for the composite optimization \eqref{eq:sorrogate}, the Fr\`echet subgradient
${\partial} F(\check{\bm{\theta}})$ is a set defined as follows \cite{rockafellar2009variational}
\begin{equation}
{\partial} F(\check{\bm{\theta}})\! :=\! \Big\{\bm{v} \, \big| \lim_{\bar{\bm \theta} \to \check{\bm{\theta}}} \inf \frac{F(\bar{\bm{\theta}})-F(\check{\bm{\theta}})-\bm{v}^\top (\bar{\bm{\theta}}-\check{\bm{\theta}}))}{\|\bar{\bm{\theta}} - \check{\bm{\theta}}\|} \ge 0 \Big\}. \nonumber
\end{equation}
Consequently, the distance between vector $\bm 0$ and the set $\partial F(\check{\bm{\theta}})$ is a measure to characterize whether a point is stationary or not.
To this aim, let us define the distance between a vector $\bm v$ and a set $\mathcal{S}$ as ${\rm dist}(\bm{v}, \mathcal{S}):= \min_{\bm{s} \in \mathcal{S}}\|\bm{v}-\bm{s}\|$. 
In this paper, we consider this notion of distance as well as the concept of $\delta$-stationary points as defined below.
\begin{definition}
	Given a small $\delta > 0$, we call vector $\check{\bm{\theta}}$ a $\delta$-stationary point if and only if
	${\rm dist}(\bm 0, \partial F(\check{\bm{\theta}})) \le \delta$.
	\label{de:stationary}
\end{definition}  

Since $f(\cdot)$ in \eqref{eq:fdef} is smooth, we have that ${\partial} F(\bar{\bm{\theta}}) = \nabla f(\bar{\bm{\theta}}) + {\partial} r(\bar{\bm{\theta}})$ \cite{rockafellar2009variational}.~Hence, it suffices to prove that the algorithm converges to a $\delta$-stationary point $\check{\bm{\theta}}$ satisfying
\begin{align}
{\rm dist}\big(\bm{0},\nabla f(\check{\bm{\theta}}) + {
\partial} r(\check{\bm{\theta}})\big) \le \delta.
\end{align}

To proceed, the following standard assumption in stochastic optimization is made here. 
\begin{assumption}
\label{as:grdestimate}
The function $f$ satisfies the next assumptions. 
\begin{enumerate}	
\item The gradient estimates are unbiased and have a bounded variance, i.e., $\mathbb{E} [  \bm g^\ast(\bar{\bm \theta}^t) - \nabla f (\bar{\bm \theta}^t) ] = \bm{0} $, and there is a constant $\sigma^2 <\infty$, so that $\mathbb{E} [\|\nabla f (\bar{\bm \theta}^t) \! - \! \bm g^\ast(\bar{\bm \theta}^t)\|_2^2] \le \sigma^2$. 
\item The function $f (\bar{\bm \theta})$ is smooth with $L_f$-Lipschitz continuous gradient, i.e., $\| \nabla f (\bar{\bm \theta}_1) - \nabla f (\bar{\bm \theta}_2)\| \le L_f \|\bar{\bm \theta}_1 - \bar{\bm \theta}_2 \|$.
\end{enumerate}
\end{assumption}

On top of Assumption \ref{as:grdestimate} and Definition \ref{de:stationary}, the next theorem provides convergence guarantees for Alg. \ref{alg:spgd}, whose proof is postponed to Appendix \ref{app:thm1}.
\begin{theorem} Let Alg. \ref{alg:spgd} run for $T$ iterations with constant step sizes $\alpha, \eta >0$. Under Assumptions \ref{as:cfunc}--\ref{as:grdestimate}, Alg. \ref{alg:spgd} generates a sequence of $\{\bm{\bar{\theta}}^t\}$ that satisfies
\begin{align}
\mathbb{E}\,[{\rm dist}(\bm{0}, \partial F(\bar{\bm \theta}^{t'}))^2] & \le  
\big(\frac{2}{\alpha} +\beta \big) \frac{\Delta_F}{T} + \big( \frac{\beta}{\eta}+2\big) \sigma^2 \nonumber \\ &\quad  + \frac{(\beta+2)L_{\bld{\bar \theta z}}^2 \epsilon}{\lambda_0}
\end{align} 
where $t'$ is uniformly sampled from $\{ 1,\ldots, T\}$; here, $\Delta_F := F(\bar{\bm{\theta}}^0) - F(\bar{\bm{\theta}}^{T+1})$; $ L_{\bld{\bar \theta} \bld  z}^2 := L_{\bld \theta \bm  z}^2 + \lambda_0 L_c$, and $\beta, \,\lambda_0>0$ are some constants.
\label{thm:convspgd}
\end{theorem}

Theorem \ref{thm:convspgd} implies the sequence $\{\bar{\bm{\theta}}^t\}_{t=1}^{T}$ generated by Alg. \ref{alg:spgd}, converges to a stationary point on average.~The upper bound here is characterized by the initial error $\Delta_F$, which decays at the rate of $\mathcal{O}(1/T)$; and, the constant bias terms induced by the gradient estimate variance $\sigma^2$ as well as the oracle accuracy~$\epsilon$.

\begin{remark}[Oracle implementation]
The $\epsilon$-accurate oracle 
can be implemented in practice by several optimization algorithms. Due to its simplicity, gradient ascent is a desirable solution.~Assuming $\gamma_{0} \ge {L_{\bm{z z}}}/{\mu}$, gradient ascent with constant step size $\eta$ obtains an $\epsilon$-accurate solution within at most $\mathcal{O}(\log({d_0^2}/{\epsilon \eta}))$ iterations, where $d_0$ is the diameter of set $\mathcal{Z}$.

The computational complexity of Alg. \ref{alg:spgd} becomes cumbersome when dealing with large-size datasets and complex models.~This motivates lightweight, scalable, yet efficient methods.~To this aim, our stochastic proximal gradient descent-ascent (SPGDA) algorithm is introduced next.

\end{remark}

\section{Stochastic Proximal Gradient Descent-Ascent}
\label{sec:spgda}
Leveraging the strong concavity of the inner maximization problem and Lemma \ref{lem:smooth}, a lightweight variant of the SPGD with $\epsilon$-accurate oracle is developed here. Instead of optimizing the inner maximization problem to $\epsilon$-accuracy by an oracle, we approximate its solution with only a \textit{single} gradient ascent step. Specifically, at every iteration $t$, for a batch of data $\{\bm{z}^t_m\}_{m=1}^{M}$, our SPGDA algorithm first perturbs each data sample via a gradient ascent step
\begin{align}
\bm{\zeta}_m^t =  \bm{z}_m^t  + \eta_t \nabla_{\bm{\zeta}} \psi(\bar{\bm \theta}^t, \bm  \zeta; \bm{z}_m^t)\big|_{\bm{\zeta}={\bm{z}_m^t}},~\forall m=1,\ldots,M \label{eq:sgaupdt}
\end{align}
then forms
\begin{align}
\bm g^{t}(\bar{\bm \theta}^t)  := \frac{1}{M} \sum_{m=1}^{M} \! \nabla_{\bar{\bm{\theta}}} \psi (\bar{\bm \theta}, \bm \zeta_m^{t}; \bm z_m^t)\big|_{\bar{\bm \theta}=\bar{\bm \theta}^{t}}. \label{eq:gsgda}
\end{align}
Having this stochastic gradient, the proposed SPGDA again takes a proximal gradient step
\begin{align}
\bar{\bm{\theta}}^{t+1} = \textrm{prox}_{\alpha_t r}
\big[ \bar{\bm{\theta}}^{t} -  \alpha_t \bm g^{t}(\bar{\bm \theta}^t) \big]. 
\label{eq:spgdaupdt}
\end{align}
The SPGDA steps are summarized in Alg. \ref{alg:spgda}. Besides its simplicity and scalability, SPGDA enjoys convergence to a stationary point as elaborated below.


\subsection{Convergence of SPGDA}
\label{sec:convspgda}

\begin{algorithm}[t]
	\SetKwInOut{Input}{Input}
	\SetKwInOut{Output}{Output}
	\SetKw{Set}{Set}	
	\Input{Initial guess $\bm{\bar{\theta}}^0$, step size sequence $\{\alpha_t, \eta_t >0 \}_{t=0}^{T}$, batch size $M$}
	
	\For{$t = 1, \ldots, T$}
	{
		{Draw a batch of i.i.d samples $\{\bm{z}_m\}_{m=1}^M$} 
		\\
		{Find $\{\bm{\zeta}_m^t\}_{m=1}^{M}$ via gradient ascent:}
		{$\bm{\zeta}_m^t =  \bm{z}_m^t  + \eta_t \nabla_{\bm{\zeta}} \psi(\bar{\bm \theta}^t, \bm  \zeta;\bm{z}_m^t)\big|_{\bm{\zeta}=\bm{z}_m^t}, \quad m = 1, \ldots, M$}
		\newline 
		{Update:  \newline $\bm{\bar{\theta}}^{t+1} = {\rm prox}_{\alpha_t r} \Big[ {\bm{\bar{\theta}}}^t-\frac{\alpha_t}{M}\sum_{m=1}^M \nabla_{\bar{\bm{\theta}}} \psi (\bm{\bar{\theta}}^t, {\bm \zeta}_m^t; \bm{z}_m^t)\big|_{\bm{\bar{\theta}}=\bm{\bar{\theta}}^t} \Big]$}
	}	  
	\caption{SPGDA}
	\label{alg:spgda}
\end{algorithm}
To establish the convergence of Alg. \ref{alg:spgda}, let
us define
\begin{align}
\bm g^\ast(\bar{\bm \theta}^t) := \frac{1}{M} \sum_{m=1}^{M} \nabla_{\bar{\bm{\theta}}} \bm \psi^\ast (\bar{\bm \theta}^{t}, \bm \zeta_m^\ast; \bm z_m^t). \label{eq:gastsgda}
\end{align}
Different from \eqref{eq:gsgda}, the gradient here is obtained
at the optimum $\bm{\zeta}_m^\ast$.~To provide the convergence result, the following assumption is made.

\begin{assumption}
The function $f$ satisfies the next assumptions. 
\begin{enumerate}
\item[1)] Gradient estimates $\nabla_{\bar{\bm{\theta}}} \bm \psi^\ast (\bar{\bm \theta}^{t}, \bm \zeta_m^\ast; \bm z_m)$ at $\bm{\zeta}_m^\ast$~are unbiased and have bounded variance. That is, for $m = 1 \cdots M$, we have $\mathbb E \, [ \nabla_{\bar{\bm{\theta}}} \psi^\ast (\bm \theta, \bm \zeta_m^\ast; \bm z_m) - \nabla_{\bar{\bm{\theta}}} f(\bm \theta)] = \bm{0}$ and  $\mathbb E \, [\|\nabla_{\bar{\bm{\theta}}} \psi^\ast (\bm \theta, \bm \zeta_m^\ast; \bm z_m)-\nabla f(\bm \theta)\|^2] \le \sigma^2$. 
\item[2)] The expected norm of $\bm{g}^t(\bar{\bm{\theta}})$ is bounded, that is, $\mathbb{E} \| \bm{g}^t(\bar{\bm{\theta}}) \|^2 \le B^2$. 
\end{enumerate}
\label{as:spgda}
\end{assumption}  
The following theorem presents convergence guarantees for Alg. \ref{alg:spgda}, whose proof is provided in Appendix \ref{thm:spgda}.  
\begin{theorem}[Convergence of Alg. \ref{alg:spgda}]
Let $\Delta_F := F(\bar{\bm{\theta}}^0) - \inf_{\bar{\bm{\theta}}}F(\bar{\bm{\theta}})$, and denote by $D$ the diameter of feasible set $\Theta$. 
Under As. \ref{as:cfunc}--\ref{as:losslipschitz} and \ref{as:spgda}, for a constant step size $\alpha > 0$, and a fixed batch size $M>0$,  after $T$ iterations, Alg. \ref{alg:spgda} satisfies 	
\begin{align}
\mathbb E  \big[   {\rm dist}&(\bm 0, {\partial} F(\bar{\bm{\theta}}^T))^2 \big] 
\le \frac{\upsilon}{T+1} \Delta_F  \nonumber \\ 
& +  \frac{2  L^2_{\bm {\theta z}} \nu}{M} \! \left[ \left(1-\alpha \mu \right) D^2 + \alpha^2 B^2 \right] + \frac{4 \sigma^2}{M}
\label{eq:convspgda}
\end{align}
where $\upsilon$, $\nu$, and $\mu = \gamma_{0} -L_{\bm{z z}}$ are some positive constants.

\label{thm:convspgda}
\end{theorem}

Theorem \ref{thm:convspgda} that implies the sequence $\{\bar{\bm \theta}^t\}_{t=1}^{T}$ generated by Alg. \ref{alg:spgda}\, converges to a stationary point. The upper bound in \eqref{eq:convspgda} is characterized by a vanishing term induced by initial error $\Delta_F$, and constant bias terms.    


\section{Distributionally Robust Federated Learning}
\label{sec:distrib}

In practice, massive datasets are distributed geographically across multiple sites, where scalability, data privacy and integrity, as well as bandwidth scarcity typically discourage uploading these data to a central server.~This has propelled the so-called federated learning framework where multiple workers exchanging information with a server to learn a centralized model using data locally generated and/or stored across workers~\cite{fedavg,fedlearn_challenges,federated_mag}.~This learning framework necessitates workers to communicate \textit{iteratively} with the server.~Albeit appealing for its scalability, one needs to carefully address the bandwidth bottleneck associated with server-worker links.~Furthermore, the workers' data may have (slightly) different underlying distributions, thus rendering the learning task challenging.~To seek a model robust to distribution shifts across workers, we capitalize on the proposed SPGDA algorithm to design a privacy- and robustness-respecting algorithm.


To that end, consider $K$ workers with each worker $k \in \mathcal{K}$ collecting samples $\{\bm{z}_n(k)\}_{n=1}^{N}$.~A globally shared model parameterized by $\bm{\theta}$ is to be updated at the server by aggregating gradients computed locally per worker.~For simplicity, here it is assumed that workers have the same number of samples $N$.~Typically, the goal is to learn a single global model from stored data at all workers by minimizing the following objective function
\begin{align}
	\underset{\bm{\theta}\in\Theta
	}{{\rm min}}\; \bar{\mathbb{E}}_{\bm{z}\sim \widehat{P}}\!\left[\ell(\bm{\theta};\bm{z}) 
	\right] + r(\bm{\theta}) 
\end{align} 
where $\bar{\mathbb{E}}_{\bm{z}\sim \widehat{P}}\!\left[\ell(\bm{\theta};\bm{z}) 
\right] := \frac{1}{NK} \sum_{n=1}^N\sum_{k=1}^{K} \ell(\bm{\theta}, \bm{z}_n(k))$.~To endow the learned model with robustness against distributional uncertainties, our novel formulation aims to solve the following problem distributedly    
\begin{align}
& \underset{\bm{\theta}\in\Theta
}{{\rm min}}\;\sup_{P\in\mathcal{P}}\;\mathbb{E}_{\bm{z}\sim P}\!\left[\ell(\bm{\theta};\bm{z}) 
\right] + r(\bm{\theta})  \nonumber \\ 
& {\textrm{s. to.}}~ \mathcal{P}:= \Big\{P \Big| \sum_{k=1}^{K} W_c(P,\widehat{P}^{(N)}(k)) \le \rho \Big\}
\label{eq:minsupdistr}
\end{align} 
where $W_c(P,\widehat{P}^{(N)}(k))$ denotes the Wasserstein distance between distribution $P$ and the locally available one $\widehat{P}^{(N)}(k)$, per worker $k$. 

Clearly, having the constraint $P \in \mathcal{P}$, couples the optimization problem in \eqref{eq:minsupdistr} across all workers.~To offer distributed implementations, we resort to Proposition \ref{prop:strongdual}, to arrive at the following equivalent reformulation 
\begin{align}
\label{eq:robustdualdistr}
\min_{\bm{\theta}\in \Theta} \, \inf_{\gamma \in \Gamma} \; \sum_{k=1}^{K}  \big\{ \bar{\mathbb{E}}_{\bm z(k) \sim \widehat{P}^{(N)}(k)} & \big[\sup_{\bm \zeta \in {\mathcal Z}} \!\left\{ \ell(\bm \theta; \bm \zeta) + \right. \\ & \left.   \gamma (\rho - c(\bm z(k), \bm \zeta))\right\} \big] \big\} + r(\bm{\theta}). \nonumber 
\end{align}

Building on the SPGDA algorithm in Section \ref{sec:spgda}, our  communication- and computation-efficient DRFL method is delineated next.
\begin{algorithm}[t]
	\SetKwInOut{Input}{Input}
	\SetKwInOut{Output}{Output}
	\SetKw{Set}{Set}	
	\Input{Initial guess $\bm{\bar{\theta}}^1$, a set of workers $\mathcal{K}$ with data samples $\{\bm{z}_n(k)\}_{n=1}^{N}$ per worker $k \in \mathcal{K}$, step size sequence $\{\alpha_t, \eta_t > 0 \}_{t=1}^{T}$ 
		
		\Output{$\bm{\bar{\theta}}^{T+1}$}	
	}
	
	\For{$t = 1, \ldots, T$}
	{
		{{\bf Each worker}:
			
			{Samples a minibatch $\mathcal{B}^t(k)$ of samples} 
			
			{Given $\bar{\bm{\theta}}^t$ and $\bm{z} \in \mathcal{B}^t(k)$, forms local perturbed loss \[\psi_k({\bar{\bm{\theta}}}^t, \bm \zeta; \bm{z}) :=  \ell(\bar{\bm{\theta}}^t; \bm \zeta) + \gamma^t (\rho - c(\bm z, \bm \zeta))\]}
			{Lazily maximizes $\psi_k({\bar{\bm{\theta}}}^t, \bm \zeta; \bm{z})$ over $\bm{\zeta}$ to find \[{\bm{\zeta}}(\bar{\bm{\theta}}^t; \bm{z}) = \bm{z} + \eta_t \nabla_{\bm{\zeta}} \psi_k (\bar{\bm{\theta}}^t, \bm{\zeta}; \bm{z})|_{\bm{\zeta}=\bm{z}}\]}
			{Computes stochastic gradient  \[\frac{1}{|\mathcal{B}^t(k)|} \sum_{\bm{z}  \in \mathcal{B}^t(k)} \!\!\! \nabla_{\bar{\bm \theta}}\psi_k (\bar{\bm{\theta}}^t, \bm{\zeta}(\bar{\bm{\theta}}^t; \bm{z}); \bm{z})\big|_{\bar{\bm{\theta}}=\bar{\bm{\theta}}^t}\] 
				and uploads to server
			}
		}

		{\bf Server}:
		
		{Updates $\bar{\bm{\theta}}^t$ according to \eqref{eq:tildthetadistr}}
		
		{Broadcasts $\bar{\bm{\theta}}^{t+1}$ to workers}
	}
	
	\caption{DRFL}
	\label{alg:drfl}
\end{algorithm}

\begin{figure*}[t]
	\centering
	\begin{subfigure}[t]{0.32\textwidth}
		\centering
		\includegraphics[width= 1 \textwidth]{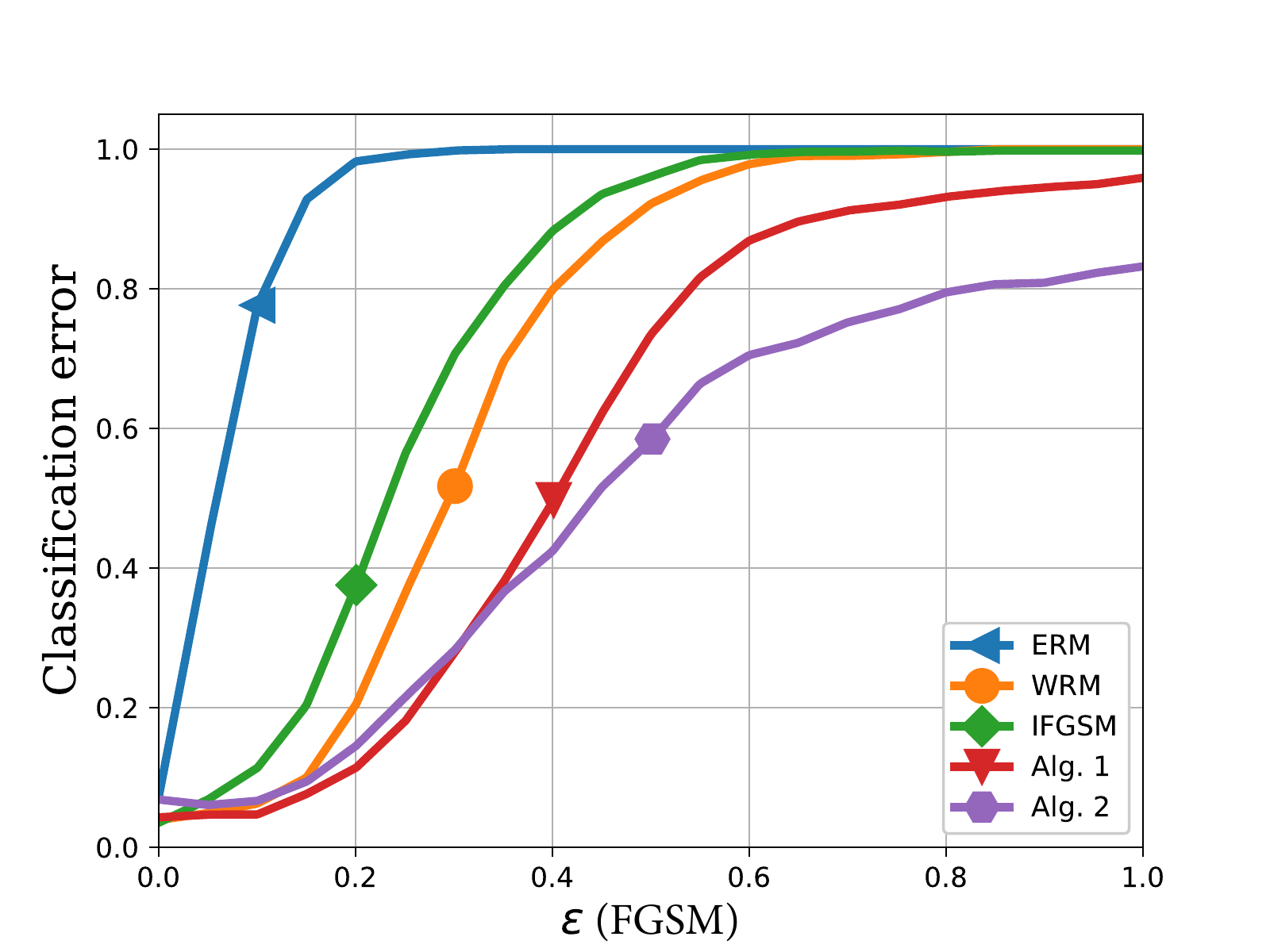}
		\caption{FGSM attack}
		\label{fig:mnsit_fgsm}
	\end{subfigure}%
	~ 
	\begin{subfigure}[t]{0.32\textwidth}
		\centering
		\includegraphics[width= 1 \textwidth]{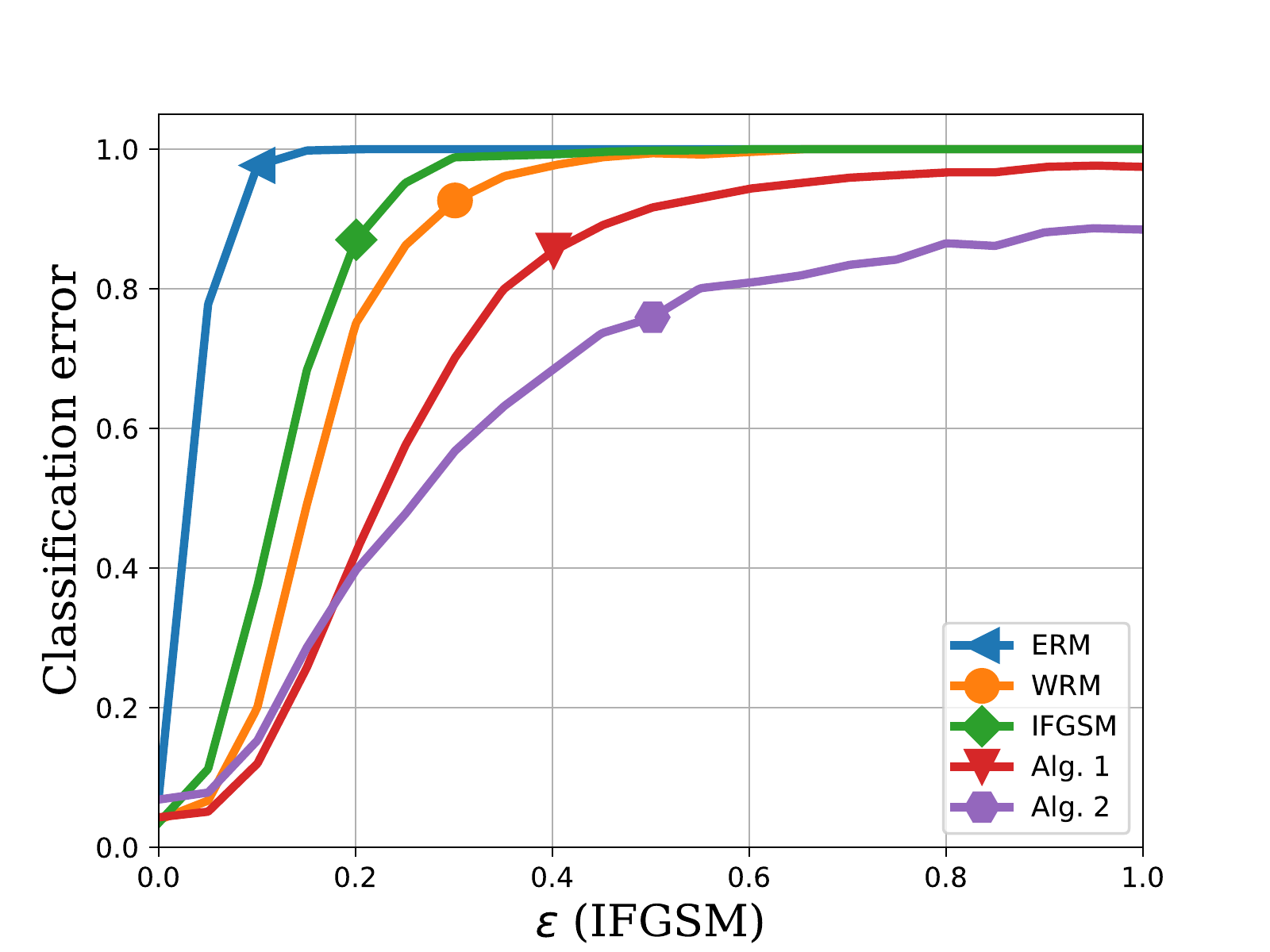}
		\caption{IFGSM attack}
		\label{fig:mnsit_ifgsm}
	\end{subfigure}%
	~
	\begin{subfigure}[t]{0.32\textwidth}
		\centering
		\includegraphics[width= 1 \textwidth]{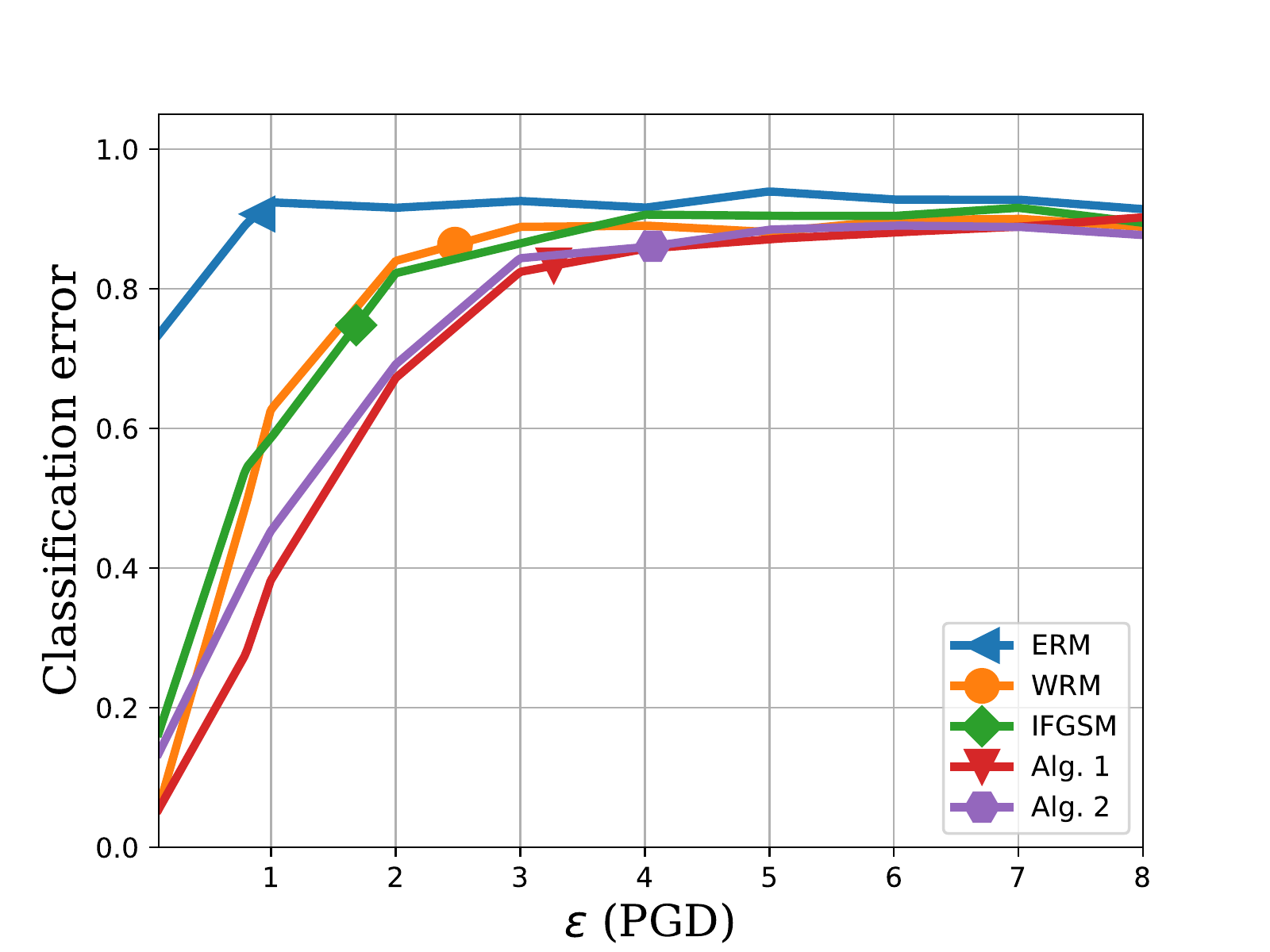}
		\caption{PGD attack}
		\label{fig:mnsit_pgd}
	\end{subfigure}%
	\caption{Misclassification error rate for different training methods using MNIST dataset.} 
	\label{fig:mnist}
\end{figure*}
\begin{figure*}[t]
	\centering\begin{subfigure}[t]{0.32\textwidth}
		\centering
		\includegraphics[width= 1 \textwidth]{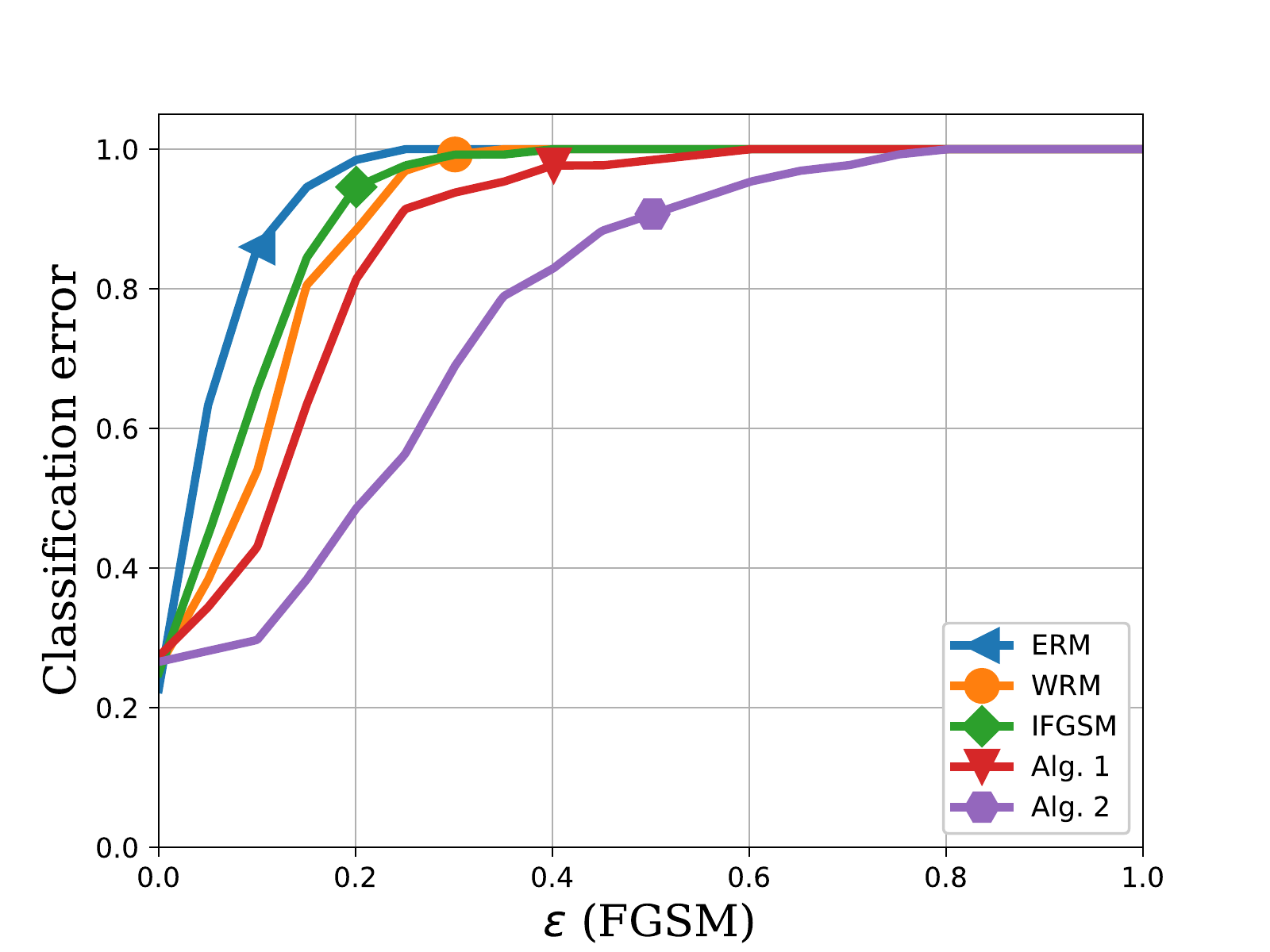}
		\caption{FGSM attack}
		\label{fig:fmnist_fgsm}
	\end{subfigure}
	~
	\begin{subfigure}[t]{0.32\textwidth}
		\centering
		\includegraphics[width= 1 \textwidth]{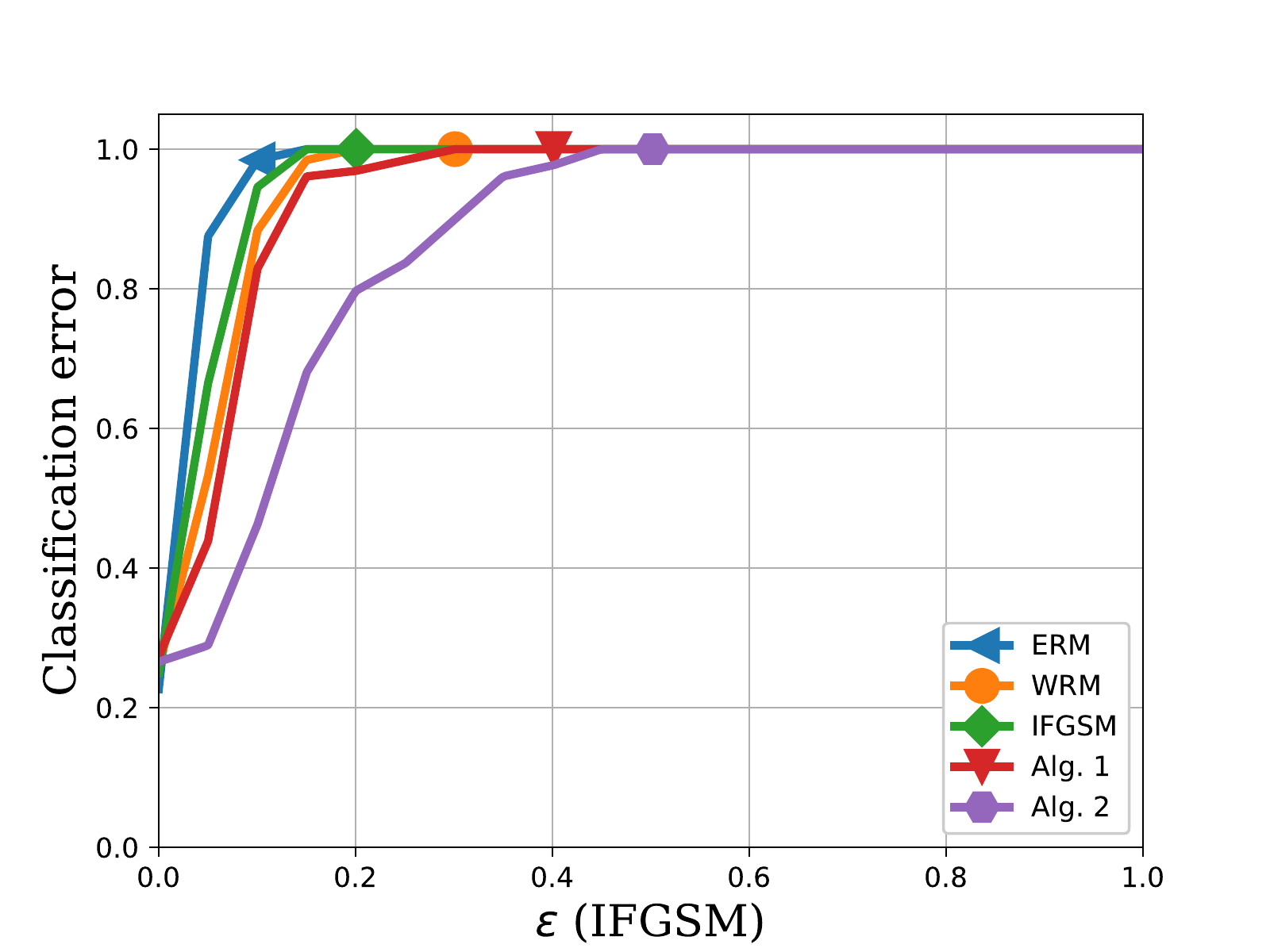}
		\caption{IFGSM attack}
		\label{fig:fmnist_ifgsm}
	\end{subfigure}
	~
	\begin{subfigure}[t]{0.32\textwidth}
		\centering
		\includegraphics[width= 1 \textwidth]{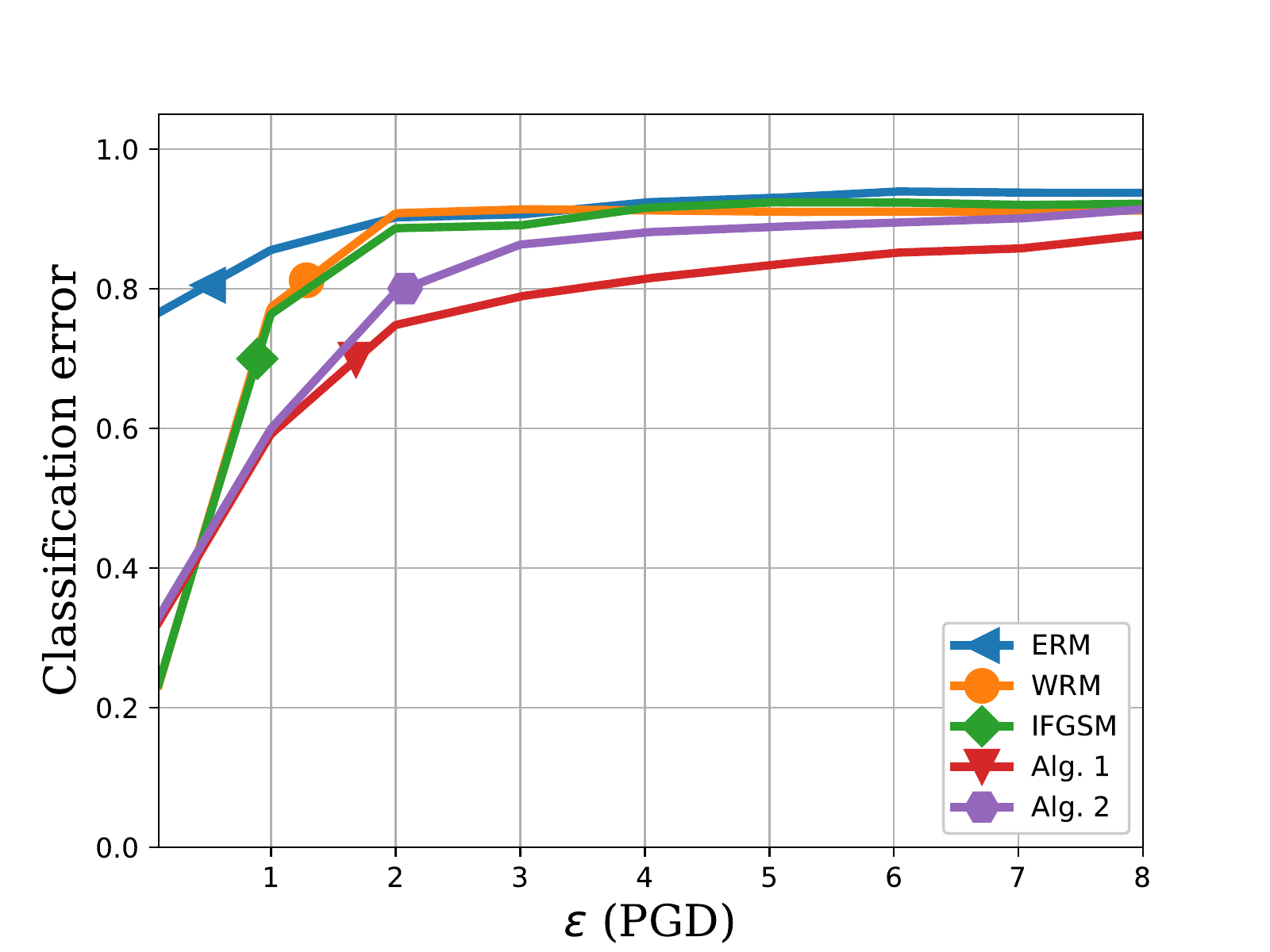}
		\caption{PGD attack}
		\label{fig:fmnist_pgd}
	\end{subfigure}
	\caption{Misclassification error rate
		for  different training methods
		using F-MNIST dataset. 
	}
	\label{fig:fmnist}
\end{figure*}
Specifically, our DRFL hinges on the fact that, at iteration $t$, having fixed the server parameters $\bar{\bm{\theta}}^t:=[{\bm{\theta}^t}^\top, \gamma^t]^\top$, the optimization problem becomes \textit{separable} across all workers. Hence, upon receiving $\bar{\bm{\theta}}^t$ from the server, each worker $k \in \mathcal{K}$: i) samples a minibatch $\mathcal{B}^t(k)$ of data from $\widehat{P}^{(N)}(k)$; ii) forms the \textit{perturbed} loss $\psi_k({\bar{\bm{\theta}}}^t, \bm \zeta; \bm{z}) :=  \ell(\bm{\theta}^t; \bm \zeta) + \gamma^t (\rho - c(\bm z, \bm \zeta))$ for each $\bm{z} \in \mathcal{B}^t(k)$; iii) lazily maximizes $\psi_k({\bar{\bm{\theta}}}^t, \bm \zeta; \bm{z})$ over $\bm{\zeta}$ using a single gradient ascent step to yield ${\bm{\zeta}}(\bar{\bm{\theta}}^t; \bm{z}) = \bm{z} + \eta_t \nabla_{\bm{\zeta}} \psi_k (\bar{\bm{\theta}}^t, \bm{\zeta}; \bm{z})|_{\bm{\zeta}=\bm{z}}$;
and, iv) sends the stochastic gradient $|\mathcal{B}^t(k)|^{-1}\sum_{\bm{z} \in \mathcal{B}^t(k)} \nabla_{\bar{\bm \theta}}\psi_k (\bar{\bm{\theta}}^t, \bm{\zeta}(\bar{\bm{\theta}}^t; \bm{z}); \bm{z})\big|_{\bar{\bm{\theta}}=\bar{\bm{\theta}}^t}$ back to the server. Upon receiving all local gradients, the server updates $\bar{\bm{\theta}}^t$ by following a proximal gradient descent step to obtain $\bar{\bm{\theta}}^{t+1}$, that is
\begin{align}
\bar{\bm{\theta}}^{t+1}   =\;   & {\rm prox}_{\alpha_t r}  \bigg[ \bar{\bm{\theta}}^t \!\! -  \frac{\alpha_t}{K}   \sum_{k=1}^{K} \frac{1}{|\mathcal{B}^t(k)|} \times \nonumber\\ & \qquad \sum_{\bm{z} \in \mathcal{B}^t(k)} \!\!\! \nabla_{\bar{\bm{\theta}}} \psi_k (\bar{\bm{\theta}}^t, \bm{\zeta}(\bar{\bm{\theta}}^t; \bm{z}); \bm{z})\big|_{\bar{\bm{\theta}}=\bar{\bm{\theta}}^t} \bigg]. \label{eq:tildthetadistr}
\end{align}
which is then broadcast to all workers to begin a new round of local updates.~Our DRFL approach is tabulated in Alg. \ref{alg:drfl}.

\section{Numerical Tests}
\label{sec:experiments}
To assess the performance against distribution shifts and adversarial perturbations, empirical evaluations on classifying standard MNIST and Fashion- (F-)MNIST datasets are presented here.~Specifically, we compare performance using models trained with empirical risk minimization (ERM), the fast-gradient method (FGSM) \cite{fellow2014adv}, its iterated variant (IFGM) \cite{ifgm2016adversarial}, and the Wasserstein robust method (WRM) \cite{sinha2017certify}. We further investigated the testing performance using the projected gradient descent (PGD) attack \cite{madry2017towards}. We start by examining the performances of SPGD with $\epsilon$-accurate oracle and the SPGDA algorithm on standard classification tasks.


\subsection{SPGD with $\epsilon$-accurate oracle and SPGDA}  
The FGSM attack performs one step gradient update along the direction of the sign of gradient to find an adversarial sample; that is,
\begin{equation}
\label{eq:fgsm}
\bm{x}_{\rm adv} = {\rm Clip}_{[-1,1]}\{\bm{x} + \epsilon_{\rm adv} {\rm sign} (\nabla \ell_{\bm{x}} (\bm{\theta}; (\bm{x}, y)))\}
\end{equation}   
where $\epsilon_{\rm avd}$ controls maximum $\ell_{\infty}$ perturbation of adversarial samples. The element-wise ${\rm Clip}_{[a, b]}\{\}$ operator enforces its input to reside in the prescribed range $[-1, 1]$. By running $T_{\rm adv}$ iterations of \eqref{eq:fgsm} iterative (I) FGSM attack samples are generated \cite{fellow2014adv}.~Starting with an initialization $\bm{x}^{0}_{\rm adv}=\bm{x}$, and considering $\ell_{\infty}$ norm, the PGD attack iterates \cite{madry2017towards}
\begin{equation}
\label{eq:ifgsm}
\bm{x}^{t+1}_{\rm adv} =  \Pi_{\mathcal{B}_{\epsilon} (\bm{x}_{\rm adv}^t)} \Big\{\bm{x}^{t}_{\rm adv} + \alpha {\rm sign} (\nabla \ell_{\bm{x}} (\bm{\theta}; (\bm{x}^{t}_{\rm adv}, y)))\Big\}
\end{equation} 
for $T_{\rm adv}$ steps, where $\Pi$ denotes projection onto the ball $\mathcal{B}_{\epsilon}(\bm{x}_{\rm adv}^t) := \{\bm{x}: \|\bm{x} - \bm{x}_{\rm adv}^t \|_\infty \le \epsilon_{\rm adv} \}$, and $\alpha>0$ is the stepsize set to $1$ in our experiments.~We use $T_{\rm adv} = 10$
iterations for all iterative methods both in training and attacks. The PGD can also be interpreted as an iterative algorithm that
solves the optimization problem $ \max_{\bm{x'}} \ell(\bm{\theta};(\bm{x}',y))$ subject to $\| \bm{x}' - \bm{x} \|_{\ell_{\infty}} \le \alpha$.~The Wasserstein attack on the other hand, generates adversarial samples through solving a perturbed  training loss with a $\ell_2$-based transportation cost associated with the Wasserstein distance between the training and adversarial data distributions~\cite{sinha2017certify}.

For the MNIST and F-MNIST datasets, a convolutional neural network (CNN) classifier consisting of $8 \times 8$, $6 \times 6$, and $5 \times 5$ filter layers with rectified linear units (ReLU) and the same padding is used.~Its first, second, and third layers have $64$, $128$, and $128$ channels, respectively, followed by a fully connected layer and a softmax layer at the output. 

CNNs with the same architecture are trained, using different adversarial samples.~Specifically, to train a Wasserstein robust CNN model (WRM), $\gamma=1$ was used to generate Wasserstein adversarial samples, $\epsilon_{\rm adv}$ was set to $0.1$
for the other two methods, and $\rho=25$ was used to define the uncertainty set for both Algs. \ref{alg:spgd} and \ref{alg:spgda}.~Unless otherwise noted, we set the batch size to $128$, the number of epochs to $30$, the learning rates to $\alpha=0.001$ and $\eta=0.02$, and used the Adam optimizer \cite{adam}.~Fig.~\ref{fig:mnsit_fgsm} shows the classification error on the MNIST dataset. The error rates were obtained using testing samples generated according to the FGSM method with $\epsilon_{\rm adv}$.~Clearly all training methods outperform ERM, and our proposed Algs. \ref{alg:spgd} and \ref{alg:spgda}  offer improved performance over competing alternatives.~The testing accuracy of all methods using samples generated according to IFGSM attack is presented in Fig.~\ref{fig:mnsit_ifgsm}.~Likewise, Algs.~\ref{alg:spgd} and \ref{alg:spgda} outperform other methods in this case.~Fig.~\ref{fig:mnsit_pgd} depicts the testing accuracy of the considered methods under different levels of PGD attack.~The plots in Fig.~\ref{fig:mnist} showcase the improved performance obtained by CNNs trained using Algs.~\ref{alg:spgd} and \ref{alg:spgda}.

%
%

%
%

\begin{figure*}[t]
	\centering
	\begin{subfigure}[t]{0.32\textwidth}
		\centering
		\includegraphics[width= 1 \textwidth]{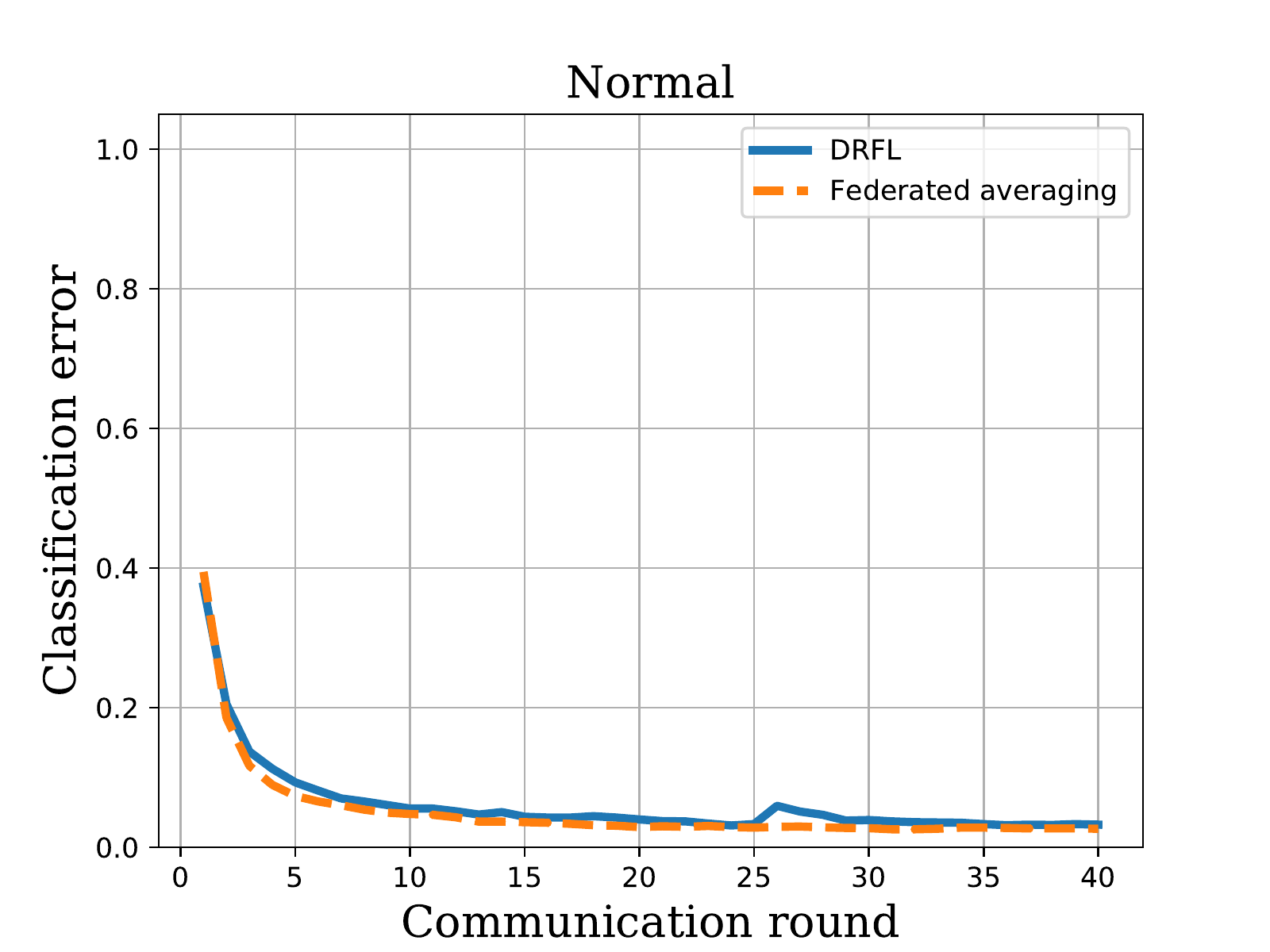}
		\caption{No attack}
		\label{fig:result_biasedfedfmnist_1}
	\end{subfigure}%
	\begin{subfigure}[t]{0.32\textwidth}
		\centering
		\includegraphics[width= 1 \textwidth]{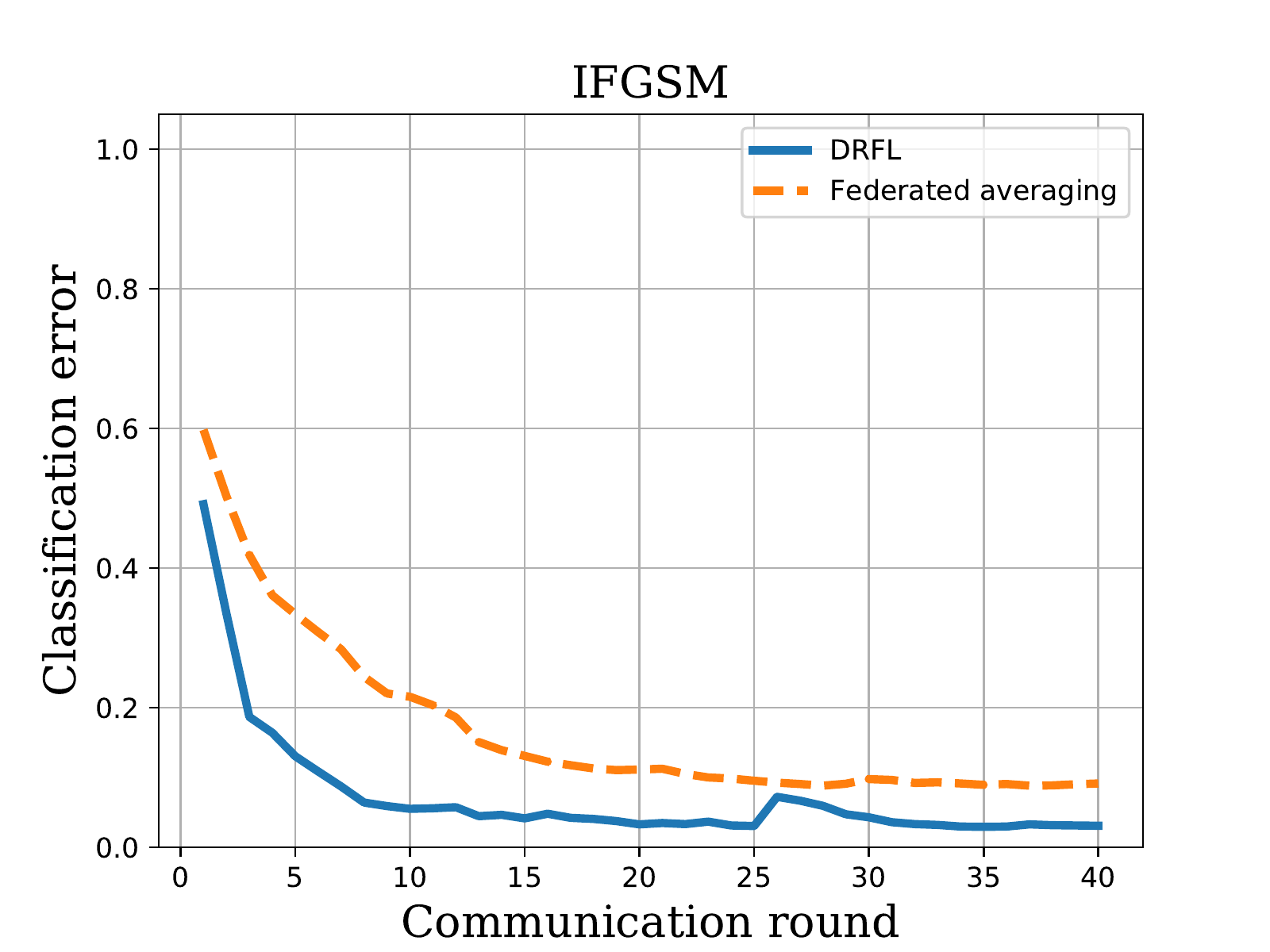}
		\caption{IFGSM attack}
		\label{fig:result_biasedfedfmnist_2}
	\end{subfigure}
	~
	\begin{subfigure}[t]{0.32\textwidth}
		\centering
		\includegraphics[width= 1 \textwidth]{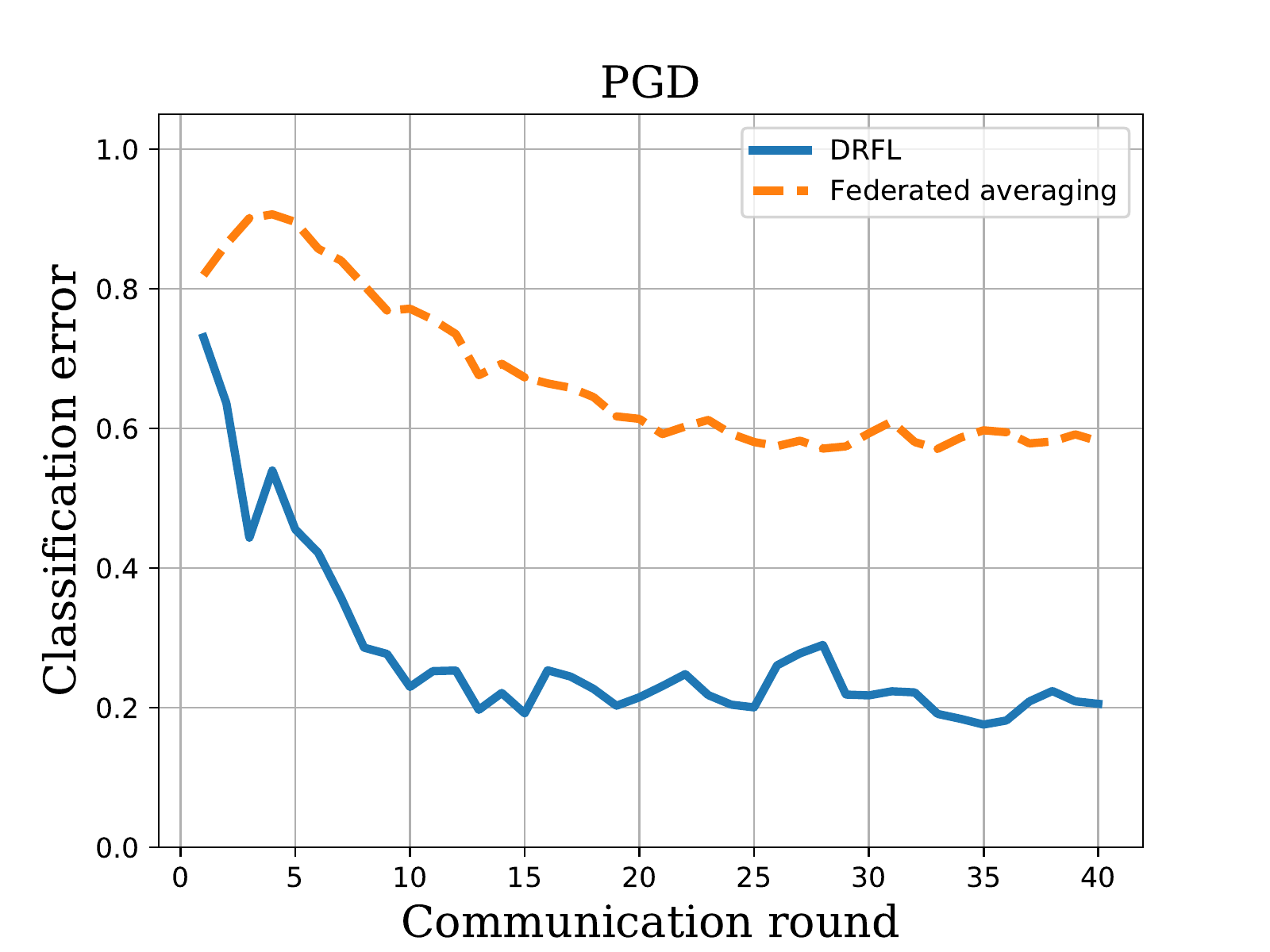}
		\caption{PGD attack}
		\label{fig:result_biasedfedfmnist_3}
	\end{subfigure}
	\caption{Distributionally robust federated learning for image classification using the non-i.i.d. F-MNIST dataset.}
	\label{fig:fedfmnist_biased}
\end{figure*}

\begin{figure*}[t]
	\centering
	\begin{subfigure}[t]{0.32\textwidth}
		\centering
		\includegraphics[width= 1 \textwidth]{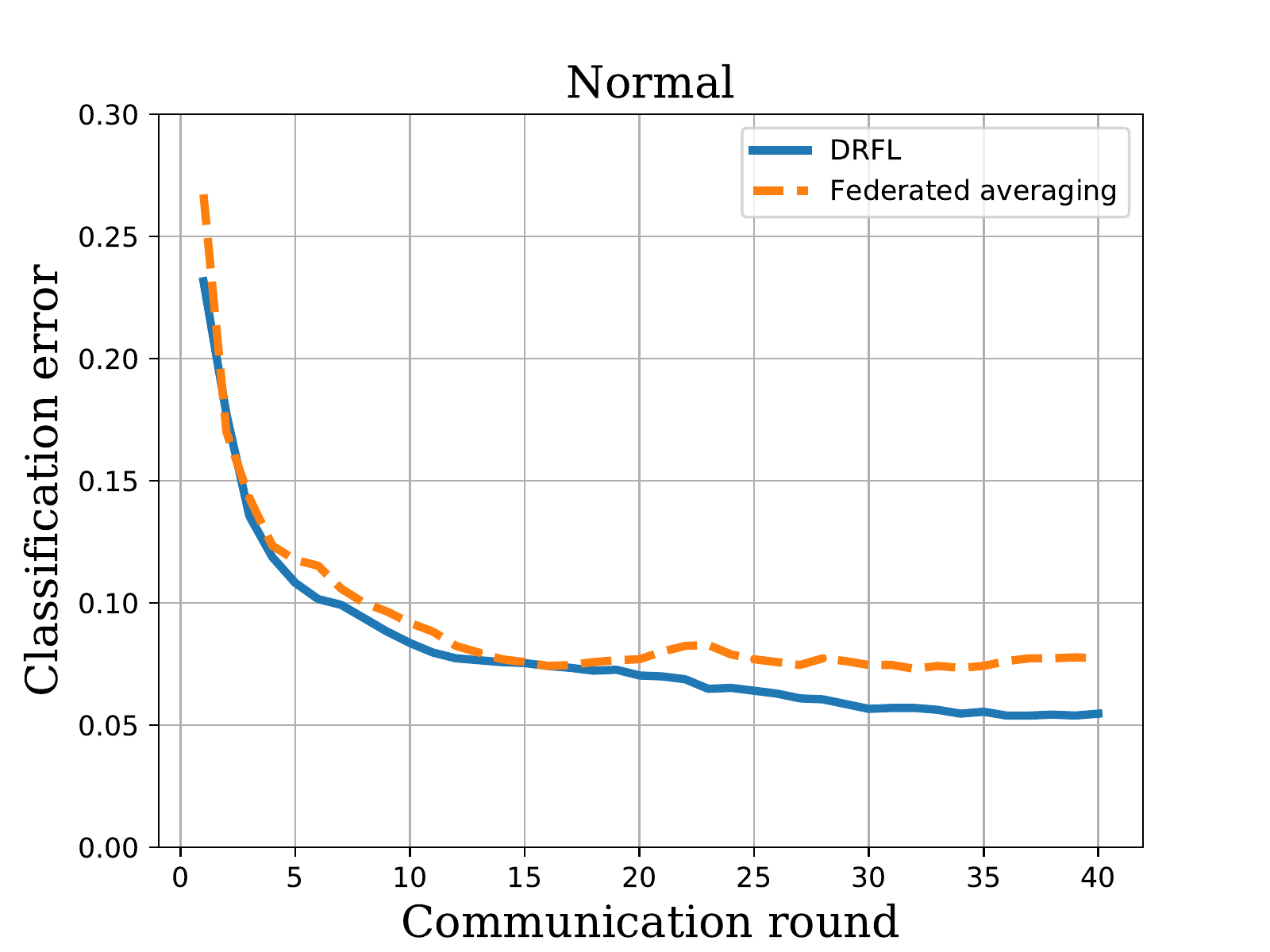}
		\caption{No attack}
		\label{fig:fedmnist_normal}
	\end{subfigure}%
	\begin{subfigure}[t]{0.32\textwidth}
		\centering
		\includegraphics[width= 1 \textwidth]{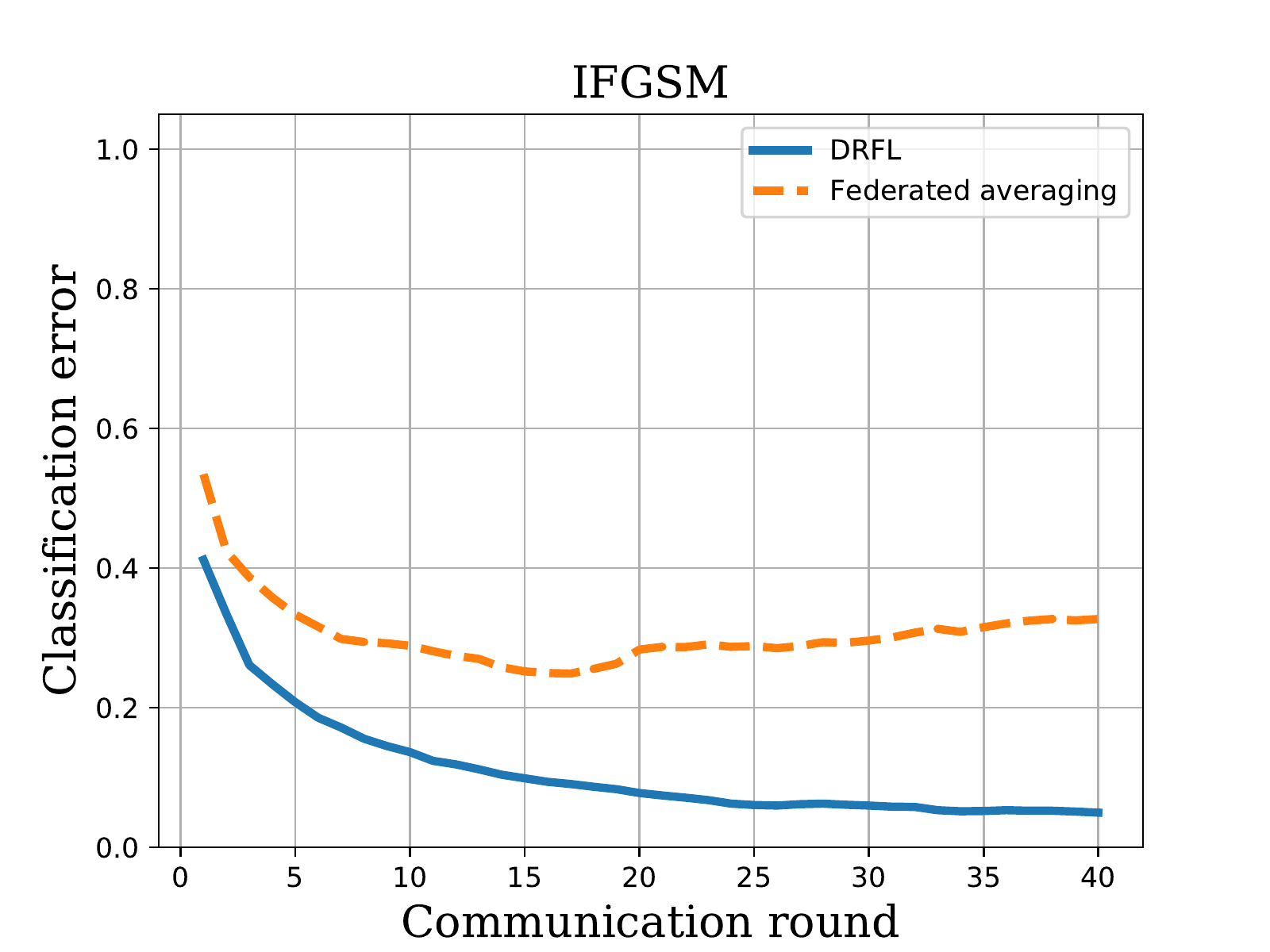}
		\caption{IFGSM attack}
		\label{fig:fedmnist_ifgsm}
	\end{subfigure}
	~
	\begin{subfigure}[t]{0.32\textwidth}
		\centering
		\includegraphics[width= 1 \textwidth]{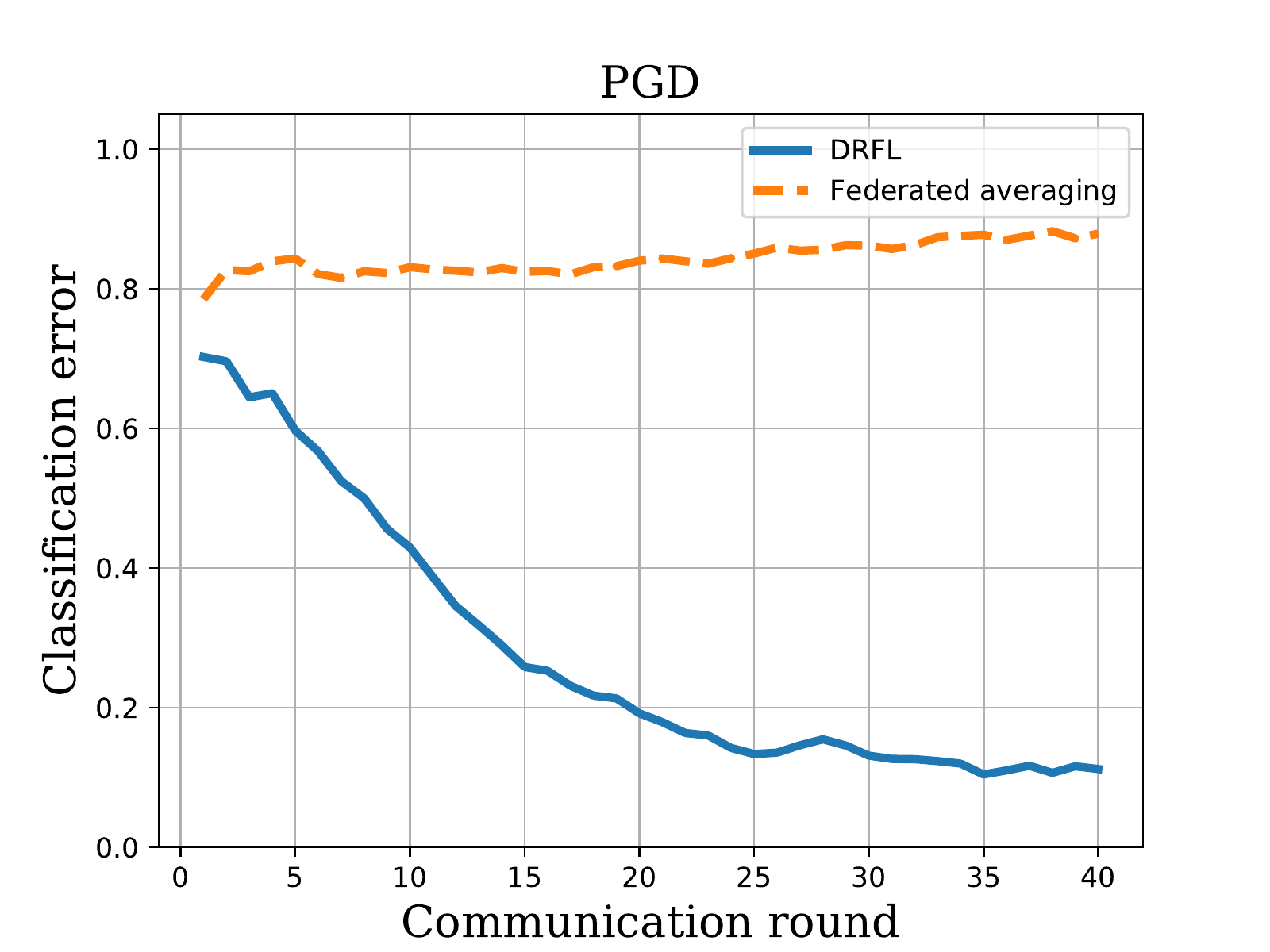}
		\caption{PGD attack}
		\label{fig:fedmnist_pgd}
	\end{subfigure}
	\caption{Federated learning for image classification using the MNIST dataset.}
	\label{fig:fedmnist}
\end{figure*}

\begin{figure*}[t]
	\centering
	\begin{subfigure}[t]{0.32\textwidth}
		\centering
		\includegraphics[width= 1 \textwidth]{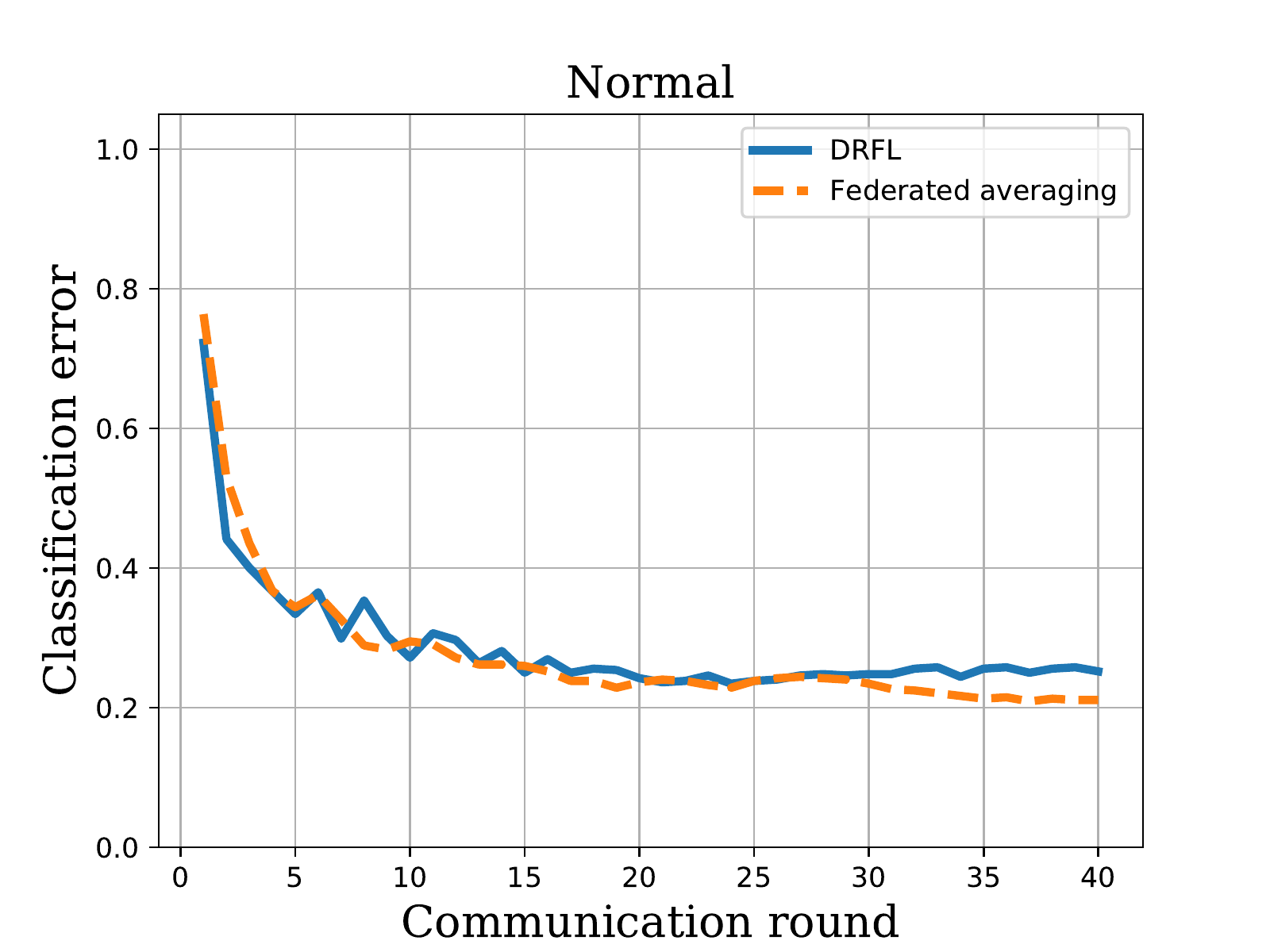}
		\caption{No attack}
		\label{fig:fedfmnist_normal}
	\end{subfigure}%
	\begin{subfigure}[t]{0.32\textwidth}
		\centering
		\includegraphics[width= 1 \textwidth]{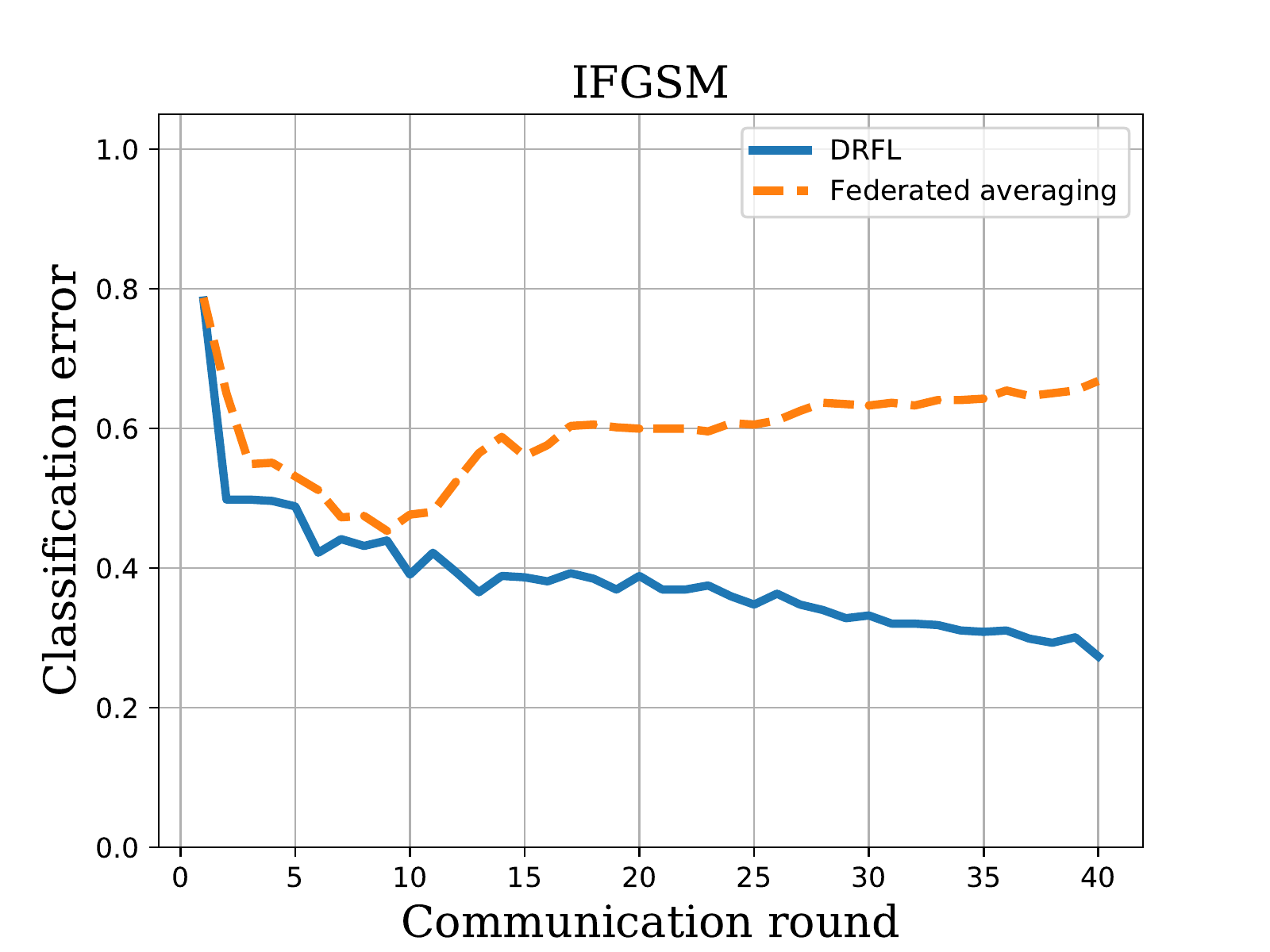}
		\caption{IFGSM attack}
		\label{fig:fedfmnist_ifgsm}
	\end{subfigure}
	~
	\begin{subfigure}[t]{0.32\textwidth}
		\centering
		\includegraphics[width= 1 \textwidth]{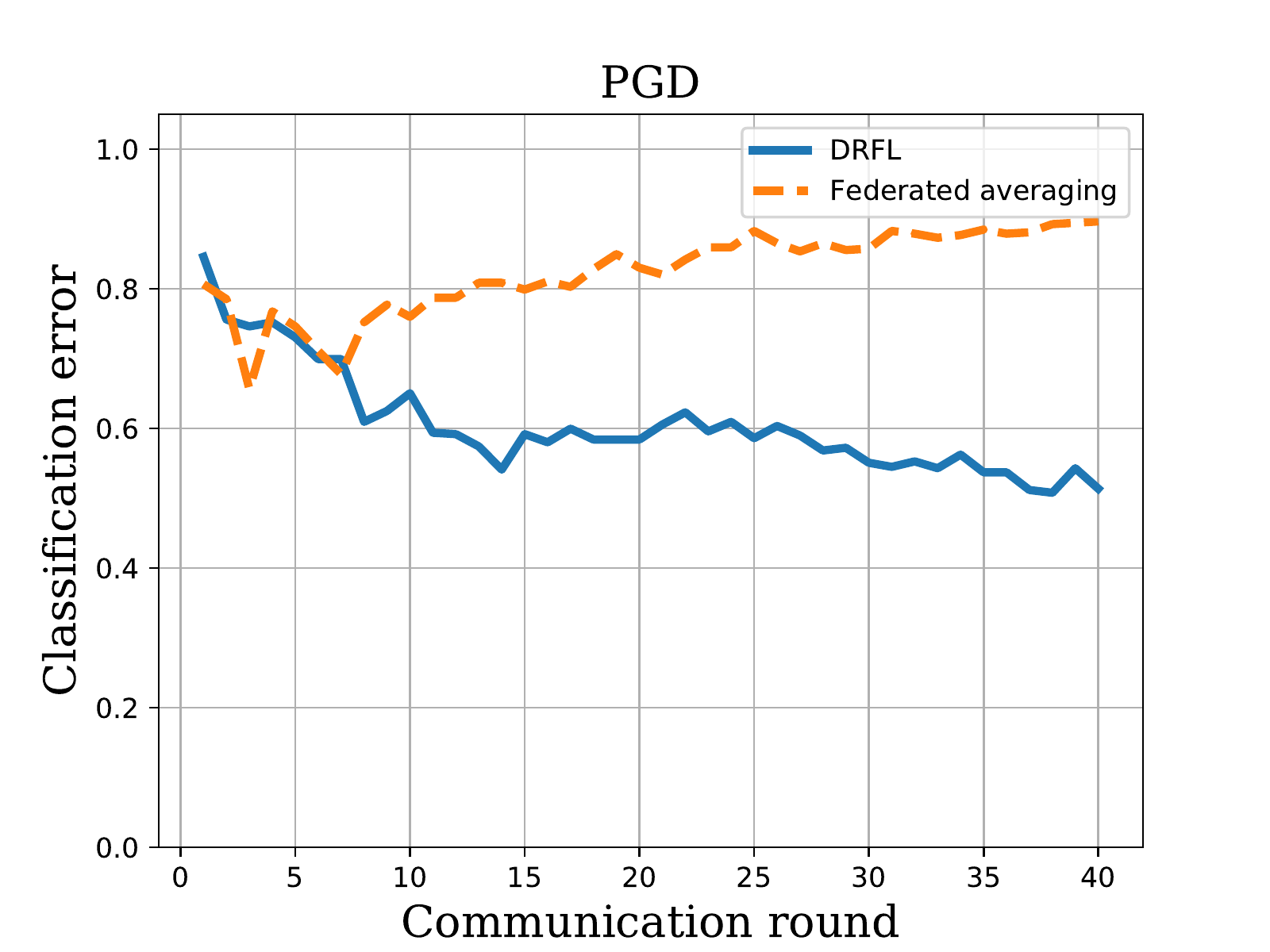}
		\caption{PGD attack}
		\label{fig:fedfmnist_pgd}
	\end{subfigure}
	\caption{Distributionally robust federated learning for image classification using F-MNIST dataset.}
	\label{fig:fedfmnist}
\end{figure*}

As for the second experiment, the F-MNIST article image dataset is adopted. Similar to MNIST dataset, each example in F-MNIST is also a $28 \times 28$ gray-scale image, associated with a label from $10$ classes.~F-MNIST is a modern replacement for the original MNIST dataset for benchmarking machine learning algorithms.~Using CNNs with similar architectures as before, the classification error is depicted for different training methods in Fig.~\ref{fig:fmnist}.  Three different attacks, namely FGSM, IFGSM, and PGD are used during testing. The resulting classification error rates are reported in Figs.~\ref{fig:fmnist_fgsm}, \ref{fig:fmnist_ifgsm}, and \ref{fig:fmnist_pgd}, respectively.
The proposed SPGD and SPGDA algorithms outperform the other training methods, verifying the superiority of Algs. \ref{alg:spgd} and \ref{alg:spgda} in terms of yielding robust models.

\subsection{Distributionally robust federated learning}
\label{sec:result_distr}
To validate the performance of our DRFL algorithm, 
we considered an FL environment consisting of a server and $10$ workers, with the local batch size $64$, and assigned to every worker is an equal-sized subset of training data containing i.i.d. samples from $10$ different classes.
In addition, it is assumed that all workers participate in each communication round.~To benchmark the DRFL, we simulated the federated averaging method \cite{fedavg}.~The testing accuracy on the MNIST dataset per communication round using clean (normal) images is depicted in Fig.~\ref{fig:fedmnist_normal}.~Clearly, both DRFL and federated averaging algorithms exhibit reasonable performance when the data is not corrupted.
The performance is further tested against IFGSM and PGD attacks with a fixed $\epsilon_{\rm adv}=0.1$ during each communication round, and the corresponding misclassification error rates are shown in Figs.~\ref{fig:fedmnist_ifgsm} and \ref{fig:fedmnist_pgd}, respectively.~The classification performance using federated averaging does not improve in Fig. \ref{fig:fedmnist_ifgsm}, whereas the DRFL keeps improving the performance across communication rounds.~This is a direct consequence of accounting for the data uncertainties during the learning process.~Furthermore, Fig.~\ref{fig:fedmnist_pgd} showcases that the federated averaging becomes even worse as the model gets progressively trained under the PGD attack.~This indeed motivates our DRFL approach when data are from untrusted entities with possibly adversarial input perturbations.~Similarly, Fig.~\ref{fig:fedfmnist} depicts the misclassification rate of the proposed DRFL method compared with federated averaging, when using F-MNIST dataset.

Since the distribution of samples across devices may influence performance, we further considered a biased local data setting.~In particular, each worker $k= 1, \ldots, 10$ contains only one class, so the data distributions at workers are highly perturbed, and data stored across workers are non-i.i.d.~The testing error rate for normal inputs is reported in Fig. \ref{fig:result_biasedfedfmnist_1}, while the test error against adversarial attacks is shown in Figs.~\ref{fig:result_biasedfedfmnist_2} and \ref{fig:result_biasedfedfmnist_3}. This set of tests shows that having distributional shifts across workers can indeed enhance testing performance when the samples are adversarially manipulated.

\section{Conclusions}
\label{sec:concls}
A framework to make parametric machine learning models robust against distributional uncertainties was proposed in this paper.~The learning task was cast as a distributionally robust optimization problem, for which two scalable stochastic optimization algorithms were developed.~The first algorithm relies on an $\epsilon$-accurate maximum-oracle to solver the inner convex subproblem, while the second approximates its solution via a single gradient ascent step.~Convergence guarantees of both algorithms to a stationary point were obtained. The upshot of the proposed approach is that it is amenable to federated learning from unreliable datasets across multiple workers. 
Our developed DRFL algorithm ensures data privacy and integrity, while offering robustness with minimal computational and communication overhead. 
Numerical tests for classifying standard real images showcased the merits of the proposed algorithms against distributional uncertainties and adversaries.~This work also opens up several interesting
directions for future research, including distributionally robust deep reinforcement learning.

\section{Appendix}
\label{appendix}

\subsection{Proof of Lemma \ref{lem:smooth}}
\label{app:lemm1}
Since function $\bm \zeta \mapsto \psi(\bar{\bm  \theta}, \bm \zeta; \bm{z})$ is $\lambda$-strongly concave, $\bm \zeta_{\ast}(\bar{\bm{\theta}}) = \sup_{\bm \zeta \in \mathcal{Z}} \psi(\bar{\bm  \theta}, \bm  \zeta; \bm{z})$ is unique. In addition, the first order optimality condition gives $\langle \nabla_{\bm \zeta} \psi(\bar{\bm{\theta}}, \bm \zeta_\ast(\bar{\bm{\theta}}); \bm z), \bm{\zeta} - \bm{\zeta}_{\ast}(\bar{\bm{\theta}})\rangle \le 0$. Let us define $\bm \zeta^1_{\ast} = \bm{\zeta}_{\ast}(\bar{\bm{\theta}}_1)$, and $\bm \zeta^2_{\ast} = \bm{\zeta}_{\ast}(\bar{\bm{\theta}}_2)$, and use strong concavity for any $\bar{\bm{\theta}}_1$ and $\bar{\bm{\theta}}_2$ to write
\begin{align}
\psi(\bar{\bm \theta}_2, \bm \zeta^2_{\ast}; \bm{z}) \le \,
& \psi(\bar{\bm \theta}_2, \bm \zeta^1_{\ast}; \bm{z}) + \langle \nabla_{\bm \zeta} \psi(\bar{\bm \theta}_2, \bm \zeta^1_{\ast}; \bm{z}), \bm \zeta^2_{\ast} - \bm \zeta^1_{\ast} \rangle \nonumber \\ 
&  -\frac{\lambda}{2} \|\bm \zeta^2_{\ast} - \bm \zeta^1_{\ast} \|^2 \label{app:lema1}
\end{align}
and
\begin{align}
\psi(\bar{\bm \theta}_2, \bm \zeta^1_{\ast}; \bm{z}) 
& \le \psi(\bar{\bm \theta}_2, \bm \zeta^2_{\ast}; \bm{z}) +  \langle \nabla_{\bm \zeta} \psi(\bar{\bm \theta}_2, \bm \zeta^2_{\ast}; \bm{z}), \bm \zeta^1_{\ast} - \bm \zeta^2_{\ast} \rangle\nonumber\\ 
& \quad -\frac{\lambda}{2} \|\bm \zeta^2_{\ast} - \bm \zeta^1_{\ast} \|^2 \nonumber \\
& \le  \psi(\bar{\bm \theta}_2, \bm \zeta^2_{\ast}; \bm{z}) -\frac{\lambda}{2} \|\bm \zeta^2_{\ast} - \bm \zeta^1_{\ast} \|^2 \label{app:lema2}
\end{align}
where the last inequality is obtained by using the first order optimality condition. Summing \eqref{app:lema1} and \eqref{app:lema2} gives
\begin{align}
&\lambda \|\bm \zeta^2_{\ast} - \bm \zeta^1_{\ast} \|^2  \le \langle \nabla_{\bm \zeta} \psi(\bar{\bm \theta}_2, \bm \zeta^1_{\ast}; \bm{z}), \bm \zeta^2_{\ast} - \bm \zeta^1_{\ast} \rangle \nonumber \\ 
& \le \langle \nabla_{\bm \zeta} \psi(\bar{\bm \theta}_2, \bm \zeta^1_{\ast}; \bm{z}), \bm \zeta^2_{\ast} - \bm \zeta^1_{\ast} \rangle -   \langle \nabla_{\bm \zeta}\psi(\bar{\bm \theta}_1, \bm \zeta^1_{\ast}; \bm{z}), \bm \zeta^2_{\ast} - \bm \zeta^1_{\ast} \rangle \nonumber \\ 
& = \langle \nabla_{\bm \zeta} \psi(\bar{\bm \theta}_2, \bm \zeta^1_{\ast}; \bm{z}) -\nabla_{\bm \zeta}\psi(\bar{\bm \theta}_1, \bm \zeta^1_{\ast}; \bm{z}) , \bm \zeta^2_{\ast} - \bm \zeta^1_{\ast} \rangle.  
\end{align}
Using H\"older's inequality we get
\begin{align}
\lambda \|\bm \zeta^2_{\ast} & - \bm \zeta^1_{\ast} \|^2  \le \nonumber \\ 
& \| \nabla_{\bm \zeta} \psi(\bar{\bm \theta}_2, \bm \zeta^1_{\ast}; \bm{z}) -\nabla_{\bm \zeta}\psi(\bar{\bm \theta}_1, \bm \zeta^1_{\ast}; \bm{z}) \|_{\star} \; \|\bm \zeta^2_{\ast} - \bm \zeta^1_{\ast} \|.
\end{align}
Therefore, 
\begin{align}
 \|\bm \zeta^2_{\ast} - \bm \zeta^1_{\ast} \| \le \frac{1}{\lambda} \| \nabla_{\bm \zeta} \psi(\bar{\bm \theta}_2, \bm \zeta^1_{\ast}; \bm{z}) -\nabla_{\bm \zeta}\psi(\bar{\bm \theta}_1, \bm \zeta^1_{\ast}; \bm{z}) \|_{\star}. \label{applem:xlimits}
\end{align}
Using $\psi(\bar{\bm{\theta}}, \bm{\zeta}; \bm{z}):= \ell(\bm{\theta}; \bm{\zeta}) + \gamma(\rho - c(\bm{z}, \bm{\zeta}))$, we have that
\begin{align}
& \| \nabla_{\bm \zeta} \psi(\bar{\bm \theta}_2, \bm \zeta^1_{\ast}; \bm{z}) -\nabla_{\bm \zeta}\psi(\bar{\bm \theta}_1, \bm \zeta^1_{\ast}; \bm{z}) \|_{\star} \nonumber \\ 
&= \| \nabla_{\bm \zeta} \ell(\bm \theta_2; \bm \zeta^1_\ast) - \nabla_{\bm{\zeta}} \ell(\bm \theta_1; \bm \zeta^1_\ast) \nonumber \\ 
& \quad + \gamma_1 \nabla_{\bm{\zeta}}  c(\bm z, \bm \zeta^1_\ast) - \gamma_2 \nabla_{\bm{\zeta}}  c(\bm z, \bm \zeta^1_\ast) \|_{\star} \nonumber \\
& \le  \| \nabla_{\bm \zeta} \ell(\bm \theta_2; \bm \zeta^1_\ast) - \nabla_{\bm{\zeta}} \ell(\bm \theta_1; \bm \zeta^1_\ast) \|_\star \nonumber \\ 
& \quad + \| \gamma_1 \nabla_{\bm{\zeta}}  c(\bm z, \bm \zeta^1_{\ast}) - \gamma_2 \nabla_{\bm{\zeta}}  c(\bm z, \bm \zeta^1_{\ast}) \|_{\star} \nonumber \\ 
& \le L_{\bm{z}\bm{\theta}} \|\bm{\theta}_2 -\bm{\bm{\theta}}_1 \| + \| \nabla_{\bm \zeta}  c(\bm z, \bm \zeta^1_{\ast}) \|_\star \, \| \gamma_2 -\gamma_1 \| \label{applem:crosslipsch}.
\end{align}
Substituting \eqref{applem:crosslipsch} into \eqref{applem:xlimits} gives
\begin{align}
\|\bm \zeta^2_{\ast} - \bm \zeta^1_{\ast} \| & \le \frac{L_{\bm{z}\bm{\theta}}}{\lambda} \|\bm{\theta}_2 -\bm{\bm{\theta}}_1 \| + \frac{1}{\lambda} \| \nabla_{\bm \zeta}  c(\bm z, \bm \zeta^1_{\ast})\|_\star \, \| \gamma_2 -\gamma_1 \|  \nonumber  \\ 
& \le \frac{L_{\bm{z}\bm{\theta}}}{\lambda}  \|\bm{\theta}_2 -\bm{\bm{\theta}}_1 \| + \frac{L_c}{\lambda}\, \| \gamma_2 -\gamma_1 \|. \label{app:lemafirtineql}
\end{align}
The last inequality holds since $\bm{\zeta} \mapsto c(\bm{z},\bm{\zeta})$ is $L_c$-Lipschitz based on the Assumption \ref{as:cstrongcvx}.  

To obtain \eqref{eq:lemmapsi}, first without loss of generality we assume that only a single datum $\bm{z}$ is given, then to prove the existence of the gradient of $\bar{\psi}(\bar{\bm{\theta}}, \bm{z})$ with respect to $\bar{\bm{\theta}}$, we resort to the Danskin's theorem as follows.  

{\bf Danskin's Theorem.} Consider the following minimax optimization problem
\begin{equation}
\min_{\bm{\theta}\in\Theta} \max_{\bm{\zeta}\in\mathcal{X}} f(\bm{\theta},\bm{\zeta}) 
\end{equation}
where $\mathcal{X}$ is a nonempty compact set, and $f: \Theta \times {\mathcal{X}} \to [0,\infty)$ is such that $f(\cdot, \bm{\zeta})$ is differentiable for any $\bm{\zeta} \in \mathcal{X}$, and $\nabla_{\bm{\theta}} f(\bm{\theta},\bm{\zeta})$ is continuous on $\Theta \times \mathcal{X}$. Let $\mathcal{S(\bm{\theta})}:=\{\bm{\zeta}_\ast| \bm{\zeta}_\ast=\arg \max_{\bm{\zeta}} f(\bm{\theta},\bm{\zeta})\}$.
Then the function 
\begin{equation*}
\bar{f}(\bm{\theta}):=\max_{\bm{\zeta}\in\mathcal{Z}} f(\bm{\theta},\bm{\zeta})
\end{equation*}
is locally Lipschitz and directionally differentiable, where the directional derivatives satisfy
\begin{equation}
\bar{f}(\bm{\theta}, \bm{d}) = \sup_{\bm{\zeta}\in \mathcal{S(\bm{\theta})}} \langle \bm{d}, \nabla_{\bm{\theta}}f(\bm{\theta},\bm{\zeta}) \rangle.
\end{equation}

For a given $\bm{\theta}$, if the set $\mathcal{S}(\bm{\theta})$ is a singleton, then the function $\bar{f}(\bm{\theta})$ is differentiable at the $\bm{\theta}$ with gradient 
\begin{equation}
\nabla_{\bm{\theta}} \bar{f}(\bm{\theta})= \nabla_{\bm{\theta}} f(\bm{\theta},\bm{\zeta}_\ast(\bm{\theta}))
\end{equation}

Given $\bm{\theta}$, and $\mu$-strongly convex function $c(\bm{z},\cdot)$, the  $\psi(\bar{\bm{\theta}}, \cdot; \bm{z})$ is concave if $L_{\bm{z z}} - \gamma {\mu} <0$, which holds true for $\gamma_0> \frac{L_{\bm{zz}}}{\mu}$.

Replacing $\bar{f}(\bm{\theta},\bm{\zeta})$ with $\psi(\bar{\bm{\theta}}, \bm \zeta; \bm{z})$, and given the concavity of $\bm \zeta \mapsto \psi(\bar{\bm{\theta}}, \bm \zeta; \bm{z})$, we have that $\bar{\psi}(\bar{\bm{\theta}}; \bm{z})$ is a continuous function with gradient
\begin{equation}
\nabla_{\bar{\bm{\theta}}} \bar{\psi}(\bar{\bm{\theta}};\bm{z})= \nabla_{\bar{\bm{\theta}}} \bar{\psi}(\bar{\bm{\theta}}, \bm{\zeta}_\ast(\bar{\bm{\theta}};\bm{z});\bm{z}).
\end{equation}

Having this result, we can obtain the second inequality, as follows
\begin{align}
& \| \nabla_{\bar{\bm{\theta}}} \psi(\bar{\bm{\theta}}_1, \bm{\zeta}^1_\ast;\bm{z}) -  \nabla_{\bar{\bm{\theta}}} \psi(\bar{\bm{\theta}}_2, \bm{\zeta}^2_\ast;\bm{z}) \| \nonumber \\ 
& \le \| \nabla_{\bar{\bm{\theta}}} \psi(\bar{\bm{\theta}}_1, \bm{\zeta}^1_\ast;\bm{z}) - \nabla_{\bar{\bm{\theta}}} \psi(\bar{\bm{\theta}}_1, \bm{\zeta}^2_\ast;\bm{z}) \|  \nonumber \\ 
& \quad + \| \nabla_{\bar{\bm{\theta}}} \psi(\bar{\bm{\theta}}_1, \bm{\zeta}^2_\ast;\bm{z}) - \nabla_{\bar{\bm{\theta}}} \psi(\bar{\bm{\theta}}_2, \bm{\zeta}^2_\ast;\bm{z}) \| \nonumber \\ 
& \le 
\left\| \begin{bmatrix}
 \nabla_{\bm{\theta}} \ell(\bm{\theta}_1,\bm{\zeta}^1_\ast) - \nabla_{\bm{\theta}} \ell(\bm{\theta}_1,\bm{\zeta}^2_\ast) \\ c(\bm{z},\bm{\zeta}^2_\ast) - c(\bm{z},\bm{\zeta}^1_\ast)
\end{bmatrix} \right\|  \nonumber \\ 
& \quad +
\left\| \begin{bmatrix}
\nabla_{\bm{\theta}} \ell(\bm{\theta}_1,\bm{\zeta}^2_\ast) - \nabla_{\bm{\theta}} \ell(\bm{\theta}_2,\bm{\zeta}^2_\ast) \\ 
0
\end{bmatrix} \right\| \nonumber \\ 
& \le L_{\bm{\theta z}} \|\bm{\zeta}^1_\ast - \bm{\zeta}^2_\ast \| + L_c \|\bm{\zeta}^1_\ast - \bm{\zeta}^2_\ast \| + L_{\bm{\theta \theta}} \|\bm{\theta_1} -\bm{\theta_2}\|  \nonumber \\ 
& \le (L_{\bm{\theta \theta}}+\frac{L_{\bm \theta z }L_{\bm{z}\bm{\theta}} + L_c L_{\bm{z}\bm{\theta}}}{\lambda})  \|\bm{\theta}_2 -\bm{\bm{\theta}}_1 \| \nonumber \\ & \quad + \frac{ L_{\bm \theta z} L_c + L^2_c}{\lambda}\, \| \gamma_2 -\gamma_1 \|
\end{align}
where we again used inequality \eqref{app:lemafirtineql}. As a technical issue, if the considered model is a neural network with a non-smooth activation function, the loss will not be continuously differentiable. However, in practice we oftentimes will not encounter these points. 


\subsection{Proof of Theorem \ref{thm:convspgd}}
\label{app:thm1}
For notational convenience, by abuse of notation let us define 
$F(\bld \theta, \gamma) := f(\bld \theta, \gamma) + r(\bld \theta) + h(\gamma) $, where $h(\gamma)$ is the indicator function defined as 
\begin{equation}
	h(\gamma) =\begin{cases}
		0, & \text{  if } \gamma \in  \Gamma \\ 
		\infty,& \text{ if } \gamma \notin  \Gamma 
	\end{cases}.
	\label{eq:gamind}
\end{equation}
where $\Gamma = \{\gamma| \gamma \ge \gamma_0\}$, and for ease of representation we use $\bar{r}(\bar{\bld \theta}) := r(\bld \theta) + h(\gamma)$. 
Having an $L_f$--smooth function $f$, yields
\begin{align} 
	f(\bld{\bar \theta}^{t+1}) &\le f(\bld{\bar \theta}^{t}) + \big \langle \nabla f(\bar{\bld \theta}^t), \bld{\bar \theta}^{t+1} - \bld{\bar \theta}^{t} \big \rangle + \frac{L_f}{2} \| \bld{\bar \theta}^{t+1} - \bld{\bar \theta}^{t} \|^2. \label{eq:smoothf1} 
\end{align}
For a given datum ${\bld z}^t$, by abuse of notation let us define the gradients
\begin{align}
\bld g^\ast(\bld {\bar \theta}^t) :=& \begin{bmatrix}
\nabla _{\bld \theta}\psi(\bld {\theta}^t, \gamma, {\bm \zeta}_\ast(\bm{\bar \theta}^t;\bm{z}^t);\bm{z}^t)
\\ 
\partial_\gamma \psi(\bld {\theta}^t, \gamma, {\bm \zeta}_\ast(\bm{\bar \theta}^t;\bm{z}^t);\bm{z}^t)
\end{bmatrix} \nonumber \\ 
= & \begin{bmatrix}
\nabla _{\bld \theta}\psi(\bld {\theta}^t, \gamma, {\bm \zeta}_\ast(\bm{\bar \theta}^t;\bm{z}^t);\bm{z}^t)
\\ 
\rho- c(\bld z^t, \bld {\bm \zeta}_\ast(\bm{\bar \theta}^t;\bm{z}^t)
\end{bmatrix}.
\label{eq:gepsdef}
\end{align}
and 
\begin{align}
\bld g^\epsilon(\bld {\bar \theta}^t) := & \begin{bmatrix}
\nabla _{\bld \theta}\psi(\bld {\theta}^t, \gamma, {\bm \zeta}_\epsilon(\bm{\bar \theta}^t;\bm{z}^t);\bm{z}^t)
\\ 
\partial_\gamma \psi(\bld {\theta}^t, \gamma, {\bm \zeta}_\epsilon(\bm{\bar \theta}^t;\bm{z}^t);\bm{z}^t)
\end{bmatrix} \nonumber \\ 
 = &
\begin{bmatrix}
\nabla _{\bld \theta}\psi(\bld {\theta}^t, \gamma, {\bm \zeta}_\epsilon(\bm{\bar \theta}^t;\bm{z}^t);\bm{z}^t)
\\ 
\rho- c(\bld z^t, \bld {\bm \zeta}_\epsilon(\bm{\bar \theta}^t;\bm{z}^t)
\end{bmatrix}
\label{eq:gepsdef}
\end{align}
obtained by an oracle at the optimal $\bm \zeta_\ast$ and the $\epsilon$-optimal $\bm \zeta_\epsilon$ solvers, respectively.  
Now, we define the error vector ${\bm \delta}(\bld{\bar \theta}^t) := \nabla f(\bld{\bar \theta}^t) -\bld g^{\epsilon}(\bld{\bar \theta}^t)$, and replace this into \eqref{eq:smoothf1} to get 
\begin{align}
	f(\bld{\bar \theta}^{t+1}) & \le f(\bld{\bar \theta}^{t}) + \big\langle \bm  g^\epsilon(\bld{\bar \theta}^t)+ \bm  \delta(\bld{\bar \theta}^t), \bld{\bar \theta}^{t+1} - \bld{\bar \theta}^{t} \big\rangle \nonumber \\ & \quad + \frac{L_f}{2} \big\| \bld{\bar \theta}^{t+1} - \bld{\bar \theta}^{t} \big\|^2. \label{eq:f_uprbnd1}
\end{align}

For proximal operator, the following  properties are equivalent for any $\bm x, \bm y$ 
\begin{align}
\label{eq:prox_prop}
\bm  u = \textrm{prox}_{\alpha r} (\bm  x)  \iff (\bm  x-\bm  u)^{T} (\bm  y - \bm  u) \le \alpha r(\bm  y) - \alpha r(\bm  u). 
\end{align}
Replacing $\bm  u=\bld{\bar \theta}^{t+1}$ and $\bm  x =\bld{\bar \theta}^t - \alpha_t \bm  g^\epsilon (\bld{\bar \theta}^t)$ in \eqref{eq:prox_prop}, we get 
\begin{equation*}
	\big\langle \bld{\bar \theta}^t - \alpha_t \bm  g^\epsilon (\bld{\bar \theta}^t) - \bld{\bar \theta}^{t+1}, \bld{\bar \theta}^t - \bld{\bar \theta}^{t+1} \big\rangle \le \alpha_t \bar{r}(\bld{\bar \theta}^t) - \alpha_t \bar{r}(\bld{\bar \theta}^{t+1}). 
\end{equation*}
Upon rearranging, we obtain 
\begin{equation}
	\big \langle \bld g^\epsilon (\bld{\bar \theta}^t), \bld{\bar \theta}^{t+1} - \bld{\bar \theta}^t \big\rangle \le \bar{r}(\bld{\bar \theta}^t) - \bar{r}(\bld{\bar \theta}^{t+1}) - \frac{1}{\alpha_t} \big\| \bld{\bar \theta}^{t+1} - \bld{\bar \theta}^{t}\big\|^2.
	\label{eq:gepsupperbound_1}
\end{equation}
Adding inequalities in \eqref{eq:gepsupperbound_1} and \eqref{eq:f_uprbnd1} gives
\begin{align*}  
	f(\bld{\bar \theta}^{t+1}) & \le f(\bld{\bar \theta}^{t}) + \big \langle \bm  \delta(\bld{\bar \theta}^t), \bld{\bar \theta}^{t+1} - \bld{\bar \theta}^{t}\big \rangle + \frac{L_f}{2} \big\| \bld{\bar \theta}^{t+1} - \bld{\bar \theta}^{t}\big\|^2  \nonumber \\ 
	& \quad + \bar{r}(\bld{\bar \theta}^t) - \bar{r}(\bld{\bar \theta}^{t+1}) - \frac{1}{\alpha_t} \big\|\bld{\bar \theta}^{t+1} - \bld{\bar \theta}^{t}\big\|^2 
\end{align*}
and we use  $F(\bm {\bar{\theta}}):= f(\bm {\bar{\theta}})+\bar{r}(\bm{\bar{\theta}})$ to obtain 
\begin{align} \nonumber 
	F(\bld{\bar \theta}^{t+1}) -  & F(\bld{\bar \theta}^{t}) \le \big \langle \bm  \delta(\bld{\bar \theta}^t), \bld{\bar \theta}^{t+1} - \bld{\bar \theta}^{t} \big\rangle \nonumber \\ 
	& \quad + \Big(\frac{L_f}{2}- \frac{1}{\alpha_t}\Big) \big\| \bld{\bar \theta}^{t+1} - \bld{\bar \theta}^{t}\big\|^2.
\end{align}
Using Young's inequality, in which for any $\eta>0$ gives $\big \langle \bm  \delta(\bld{\bar \theta}^t), \bld{\bar \theta}^{t+1} - \bld{\bar \theta}^{t} \big\rangle \le \frac{\eta}{2} \| \bld{\bar \theta}^{t+1} - \bld{\bar \theta}^{t}\|^2 +   \frac{1}{2\eta} \| \delta(\bld{\bar \theta}^t)\|^2$, yielding    
\begin{align} 
	F(\bld{\bar \theta}^{t+1}) -  & F(\bld{\bar \theta}^{t}) \le  \Big(\frac{L_f + \eta}{2}\!-\! \frac{1}{\alpha_t}\Big) \big\|\bld{\bar \theta}^{t+1} \!-\! \bld{\bar \theta}^{t}\big\|^2 + \frac{\big\|\bm  \delta(\bld{\bar \theta}^t)\big\|^2}{2 \eta}
	\label{eq:bnd_itrsF1}.
\end{align}

Next we are going to bound $\bm\delta(\bld{\bar \theta}^t) := \nabla f (\bld{\bar \theta}^t) - \bm g^\epsilon(\bld{\bar \theta}^t)$. By adding and subtracting $\bld g^\ast(\bld{\bar \theta}^t)$ to the right hand side, we get
\begin{equation}
	\label{eq:bounddelta1}
	\|\bm\delta(\bld{\bar \theta}^t)\|^2 \le 2 \|\nabla f (\bld{\bar \theta}^t) - \bm g^\ast(\bld{\bar \theta}^t)\|^2 + 2 \|\bm g^\ast(\bld{\bar \theta}^t) - \bm g^\epsilon(\bld{\bar \theta}^t)\|^2. 
\end{equation}

Due to the Lipschitz smoothness of the gradient, it holds that 
\begin{align}
	& \big\|\bm g^\ast(\bld{\bar \theta}^t) - \bm g^\epsilon(\bld{\bar \theta}^t)\big\|^2 
	\nonumber \\ 
	& = \Bigg\| \begin{bmatrix}
	\nabla _{\bld \theta}\psi(\bld {\theta}^t, \gamma, {\bm \zeta}_\ast(\bm{\bar \theta}^t;\bm{z}^t);\bm{z}^t)
	\\ 
	\rho- c(\bld z^t, \bld {\bm \zeta}_\ast(\bm{\bar \theta}^t;\bm{z}^t)
	\end{bmatrix}  
	\nonumber \\ 
	& \quad - \begin{bmatrix}
	\nabla _{\bld \theta}\psi(\bld {\theta}^t, \gamma, {\bm \zeta}_\epsilon(\bm{\bar \theta}^t;\bm{z}^t);\bm{z}^t)
	\\ 
	\rho- c(\bld z^t, \bld {\bm \zeta}_\epsilon(\bm{\bar \theta}^t;\bm{z}^t)
	\end{bmatrix} \Bigg\|^2 \nonumber \\ 
& = \big \|\nabla _{\bld \theta}\psi(\bld {\theta}^t, \gamma, {\bm \zeta}_\ast(\bm{\bar \theta}^t;\bm{z}^t);\bm{z}^t)- \nabla _{\bld \theta}\psi(\bld {\theta}^t, \gamma, {\bm \zeta}_\epsilon(\bm{\bar \theta}^t;\bm{z}^t);\bm{z}^t) \big \|^2  \nonumber \\ 
& \quad  + \big \| c(\bld z^t, \bld \zeta^t_\ast) - c(\bld z^t,\bld \zeta^t_\epsilon) \big \|^2 \nonumber \\ 
& \overset{(\rm a)}{\le} \Big(\frac{ L_{\bld \theta \bm  z}^2  }{\lambda^t}+L_c\Big) \|\bld \zeta^t_\ast-\bld \zeta^t_\epsilon \|^2  \nonumber \\ 
& \overset{(\rm b)}{\le}  \Big(\frac{ L_{\bld \theta \bm  z}^2  }{\lambda^t}+L_c \Big) \epsilon \nonumber \\
& \le \Big(\frac{ L_{\bld \theta \bm  z}^2  }{\lambda_{0}}+L_c \Big)\epsilon \label{eq:boundg1}
\end{align}
where $(\rm a)$ uses the $\lambda^t = \mu \gamma^t -L_{\bld z \bld z}$ strong-concavity of ${\bm \zeta}\mapsto \psi(\bar{\bld \theta}, \gamma, \bm{\zeta}; \bld z)$, and the second term is bounded by $L_c \|\bld \zeta_{\ast}^t - \bld \zeta_{\epsilon}^t\|^2$ according to Assumption \ref{as:cstrongcvx}. The last inequality holds for $\lambda_{0}:= \mu \gamma_0 - L_{\bm{zz}}$, where we used \eqref{eq:gamind} to bound $\gamma^t \ge \gamma_{0} > L_{\bld z\bld z}$. So far, we have established the following 
\begin{equation}
	\|\bm g^\ast(\bld{\bar \theta}^t) - \bm g^\epsilon(\bld{\bar \theta}^t)\|^2 \le \frac{ L_{\bld{\bar \theta} \bm  z}^2  \epsilon}{\lambda_0}
	\label{eq:ggbound}
\end{equation}
where for notational convenience we used  $ L_{\bld{\bar \theta} \bld  z}^2   = L_{\bld \theta \bm  z}^2 + \lambda_0 L_c$. Upon replacing \eqref{eq:ggbound} into \eqref{eq:bounddelta1}, we bound the error as follows 
\begin{equation}
	\label{eq:bound_delta2}
	\|\bm\delta(\bld{\bar \theta}^t)\|^2 \le 2 \| \nabla f (\bld{\bar \theta}^t) - \bm g^\ast(\bld{\bar \theta}^t)\|^2 + \frac{ 2 L_{\bld{\bar \theta} \bm  z}^2  \epsilon}{\lambda_0}.
\end{equation}
Combining \eqref{eq:bnd_itrsF1} and \eqref{eq:boundg1} yields
\begin{align} \nonumber 
	F(\bld{\bar \theta}^{t+1}) -  F(\bld{\bar \theta}^{t}) & \le \Big( \frac{L_f+\eta}{2}- \frac{1}{\alpha_t}\Big) \big\|\bld{\bar \theta}^{t+1} - \bld{\bar \theta}^{t}\big\|^2 \\
	& \quad +  \frac{1}{\eta} \big\|\nabla f (\bld{\bar \theta}^t) - \bm g^\ast(\bld{\bar \theta}^t)\big\|^2 + \frac{ L_{\bld{\bar \theta} \bm  z}^2  \epsilon}{\eta \lambda_0}.
\end{align}
Considering a constant step size $\alpha$ and summing these inequalities over $t= 1, \ldots, T$ yields
\begin{align} 
	\Big( \frac{1}{\alpha} - & \frac{L_f + \eta}{2}\Big) \sum_{t = 0}^{T} \big\|\bld{\bar \theta}^{t+1} - \bld{\bar \theta}^{t}\big\|^2 \le F(\bld{\bar \theta}^{0}) -  F(\bld{\bar \theta}^{T})  +  \nonumber \\
	& \quad  \frac{1}{\eta} \sum_{t=0}^{T} \big\|\nabla f (\bld{\bar \theta}^t) - \bm g^\ast(\bld{\bar \theta}^t)\big\|^2 + \frac{ (T+1) L_{\bld{\bar \theta} \bm  z}^2  \epsilon}{\lambda_0}.
	\label{eq:boundtheta1}
\end{align}
From proximal gradient update
\begin{equation}
	\bld{\bar \theta}^{t+1} = \arg \min_{\bld \theta} \, \alpha \bar{r}(\bld \theta) + \alpha \big\langle \bld \theta - \bld{\bar \theta}^t, \bm g^\epsilon (\bld{\bar \theta}^t) \big \rangle + \frac{1}{2} \big\|\bld \theta - \bld{\bar \theta}^{t}\big\|^2 \label{eqapp:proxupdate}
\end{equation}
clearly, due to optimality of $\bld{\bar \theta}^{t+1}$ in \eqref{eqapp:proxupdate}, it holds that
\begin{equation}
	\bar{r}(\bld{\bar \theta}^{t+1}) + \big\langle \bld{\bar \theta}^{t+1} - \bld{\bar \theta}^t, \bm g^\epsilon (\bld{\bar \theta}^t) \big\rangle + \frac{1}{2 \alpha} \big\|\bld{\bar \theta}^{t+1} - \bld{\bar \theta}^{t}\big\|^2 \le \bar{r}(\bld{\bar \theta}^{t})
\end{equation}
which combined with the smoothness of $f$ (c.f. \eqref{eq:smoothf1}) yields
\begin{align}
	\nonumber
\big\langle \bld{\bar \theta}^{t+1} - \bld{\bar \theta}^t , \bm g^\epsilon (\bld{\bar \theta}^t) - \nabla f(\bld{\bar \theta}^t) \big\rangle + &  \Big(\frac{1}{2 \alpha} - \frac{L_f}{2}\Big) \big\|\bld{\bar \theta}^{t+1} - \bld{\bar \theta}^{t}\big\|^2 \nonumber \\
	& \le F(\bld{\bar \theta}^{t}) - F(\bld{\bar \theta}^{t+1}) 
\end{align}
Subtracting $\langle \bld{\bar \theta}^{t+1} - \bld{\bar \theta}^t, \nabla f(\bld{\bar \theta}^{t+1}) \rangle$ from both sides gives
\begin{align*}
\big\langle \bld{\bar \theta}^{t+1} -   
	\bld{\bar \theta}^t),  \bm g^\epsilon (\bld{\bar \theta}^t)  - \nabla f(\bld{\bar \theta}^{t+1}) \big \rangle 
	+ \Big(\frac{1}{2 \alpha}  -\frac{L_f}{2}\Big) \big\|\bld{\bar \theta}^{t+1} - \bld{\bar \theta}^{t}\big\|^2 \nonumber  \\ 
	\le F(\bld{\bar \theta}^{t}) - F(\bld{\bar \theta}^{t+1}) -  \big\langle  \bld{\bar \theta}^{t+1} - \bld{\bar \theta}^t, \nabla f (\bld{\bar \theta}^{t+1}) - \nabla f(\bld{\bar \theta}^t) \big\rangle.
\end{align*}
By forming $\big\|\bm g^\epsilon (\bld{\bar \theta}^t) - \nabla f(\bld{\bar \theta}^{t+1}) + \frac{1}{\alpha} (\bld{\bar \theta}^{t+1} - \bld{\bar \theta}^t)\big\|^2$ on the left hand side and adding relevant terms to the right hand side, we arrive at
\begin{align}
	& \Big\|\bm g^\epsilon (\bld{\bar \theta}^t) - \nabla f(\bld{\bar \theta}^{t+1}) + \frac{1}{\alpha} (\bld{\bar \theta}^{t+1} - \bld{\bar \theta}^t)\Big\|^2 \nonumber \\
	&  \le  \big\|\bm g^\epsilon (\bld{\bar \theta}^t) - \nabla f(\bld{\bar \theta}^{t+1})\big\|^2 
	+ \frac{1}{\alpha^2} \big\|\bld{\bar \theta}^{t+1} - \bld{\bar \theta}^t\big\|^2  \nonumber \\
	& \quad + \Big(\frac{L_f}{\alpha} - \frac{1}{\alpha^2}\Big)\big\|\bld{\bar \theta}^{t+1} - \bld{\bar \theta}^{t}\big\|^2 + \frac{2}{\alpha}\big(F(\bld{\bar \theta}^{t}) - F(\bld{\bar \theta}^{t+1})\big)
	\nonumber \\ 
	& \quad   - \frac{2}{\alpha}\big\langle  \bld{\bar \theta}^{t+1} - \bld{\bar \theta}^t, \nabla f (\bld{\bar \theta}^{t+1}) - \nabla f(\bld{\bar \theta}^t) \big\rangle  \nonumber \\ 
	& \le  \big\|\bm g^\epsilon (\bld{\bar \theta}^t) - \nabla f(\bld{\bar \theta}^{t})\|^2 + \frac{1}{\alpha^2} \big\|\bld{\bar \theta}^{t+1} - \bld{\bar \theta}^t\big\|^2 \nonumber \\
	& \quad  + (\frac{L_f}{\alpha} - \frac{1}{\alpha^2}) \big\|\bld{\bar \theta}^{t+1} - \bld{\bar \theta}^{t}\big\|^2 + \frac{2}{\alpha}\big(F(\bld{\bar \theta}^{t}) - F(\bld{\bar \theta}^{t+1})\big) \nonumber \\ 
	& \quad   - \frac{2}{\alpha} \big\langle \bld{\bar \theta}^{t+1} - \bld{\bar \theta}^t), \nabla f (\bld{\bar \theta}^{t+1}) - \nabla f(\bld{\bar \theta}^t) \big \rangle \nonumber \\
	& \le \big\|\bm g^\epsilon (\bld{\bar \theta}^t) - \nabla f(\bld{\bar \theta}^{t})\|^2  + \frac{1}{\alpha^2} \big\|\bld{\bar \theta}^{t+1} - \bld{\bar \theta}^{t}\big\|^2 \nonumber \\ 
	& \quad + \Big(\frac{L_f}{\alpha} - \frac{1}{\alpha^2}\Big) \big\|\bld{\bar \theta}^{t+1} - \bld{\bar \theta}^{t}\big\|^2 + \frac{2}{\alpha}\big(F(\bld{\bar \theta}^{t}) - F(\bld{\bar \theta}^{t+1})\big) \nonumber \\ 
	& \quad + \frac{\eta}{\alpha} \big\|\bld{\bar \theta}^{t+1} - \bld{\bar \theta}^{t}\big\|^2 +\frac{L_f^2}{\eta} \big\|\bld{\bar \theta}^{t+1} - \bld{\bar \theta}^{t}\big\|^2.
\end{align}
Here the last inequality is obtained by applying Young's inequality and then using the $L_f$-Lipschitz continuity of function~$f(\cdot)$. By simplifying the last inequality we get
\begin{align}
	 \Big\|\bm g^\epsilon (\bld{\bar \theta}^t) & - \nabla f(\bld{\bar \theta}^{t+1}) + \frac{1}{\alpha} (\bld{\bar \theta}^{t+1}  - \bld{\bar \theta}^t)\Big\|^2 \nonumber \\ 
	& \le \big\|\bm g^\epsilon (\bld{\bar \theta}^t) - \nabla f(\bld{\bar \theta}^{t})\big\|^2  + \frac{2}{\alpha}\big(F(\bld{\bar \theta}^{t}) - F(\bld{\bar \theta}^{t+1})\big) \nonumber  \\
	& \quad   + \Big(\frac{L_f^2}{\eta} + \frac{L_f+\eta}{\alpha} \Big) \big\|\bld{\bar \theta}^{t+1} - \bld{\bar \theta}^{t}\big\|^2.
\end{align}
The first term in the right hand side can be bounded by adding and subtracting $\bm{g}^\ast(\bar{\bm{\theta}}^t)$ and using \eqref{eq:ggbound}, to arrive at 
\begin{align}
	\Big\|\bm g^\epsilon  & (\bld{\bar \theta}^t) - \nabla f(\bld{\bar \theta}^{t+1}) + \frac{1}{\alpha} (\bld{\bar \theta}^{t+1}   - \bld{\bar \theta}^t)\Big\|^2 \nonumber \\ & \le 2 \big\|\nabla f (\bld{\bar \theta}^t) - \bm g^\ast(\bld{\bar \theta}^t)\big\|^2     \frac{2 L_{\bld{\bar \theta}}^2 \epsilon }{\lambda_0}  +  \frac{2}{\alpha}\big(F(\bld{\bar \theta}^{t}) - F(\bld{\bar \theta}^{t+1})\big) \nonumber \\ 
	& \quad + \Big(\frac{L_f^2}{\eta} + \frac{L_f+\eta}{\alpha} \Big) \big\|\bld{\bar \theta}^{t+1} - \bld{\bar \theta}^{t}\big\|^2. \label{eq:subgradientbound}
\end{align}
Sum these inequalities over $t=1, \ldots, T$, to get
\begin{align}
	& \sum_{t=0}^{T} \Big\|\bm g^\epsilon (\bld{\bar \theta}^t) - \nabla f(\bld{\bar \theta}^{t+1}) + \frac{1}{\alpha} (\bld{\bar \theta}^{t+1} - \bld{\bar \theta}^t)\Big\|^2 \nonumber \\ 
	& \le 2 \sum_{t=0}^{T} \big\|\nabla f (\bld{\bar \theta}^t) - \bm g^\ast(\bld{\bar \theta}^t)\big\|^2 +  \frac{2 (T+1) L_{\bld{\bar \theta}}^2 \epsilon }{\lambda_0} \nonumber \\  
	& \quad + \frac{2}{\alpha}\big(F(\bld{\bar \theta}^{0}) - F(\bld{\bar \theta}^{T})\big) \nonumber \\ 
	& \quad + \Big(\frac{L_f^2}{\eta} + \frac{L_f+\eta}{\alpha} \Big) \sum_{t=0}^{T} \big\|\bld{\bar \theta}^{t+1} - \bld{\bar \theta}^{t}\big\|^2. 
\end{align}
Using \eqref{eq:boundtheta1} to bound the last term yields  
\begin{align}
	\sum_{t=0}^{T} &  \Big\|\bm g^\epsilon (\bld{\bar \theta}^t) - \nabla f(\bld{\bar \theta}^{t+1}) + \frac{1}{\alpha} (\bld{\bar \theta}^{t+1} - \bld{\bar \theta}^t)\Big\|^2 \nonumber \\ 
	& \le 2 \sum_{t=0}^{T} \left\|\nabla f (\bld{\bar \theta}^t) - \bm g^\ast(\bld{\bar \theta}^t)\right\|^2 + \frac{2 (T+1) L_{\bld{\bar \theta}}^2 \epsilon }{\lambda_0}  \nonumber \\ 
	& \quad + \frac{2}{\alpha} \Delta_F  + \beta \Delta_F + \frac{\beta}{\eta} \sum_{t=0}^{T} \big\|\nabla f (\bld{\bar \theta}^t) - \bm g^\ast(\bld{\bar \theta}^t)\big\|^2 \nonumber \\ 
	& \quad + \frac{\beta (T+1)L^2_{\bld{\bar \theta}\bld z}\epsilon}{\lambda_0} 
\end{align}
where $\beta = (\frac{L_f^2}{\eta} + \frac{L_f+\eta}{\alpha}) \frac{2 \alpha}{2-(L_f+\eta)\alpha}$. By taking expectation of both sides of this inequality, we obtain
\begin{align}
\label{eq:bnd_sbdif}
	\frac{1}{T+1}\mathbb E \Big[ \sum_{t=0}^{T} \Big\|\bm g^\epsilon (\bld{\bar \theta}^t)  - \nabla f(\bld{\bar \theta}^{t+1}) \! + \! \frac{1}{\alpha} (\bld{\bar \theta}^{t+1} - \bld{\bar \theta}^t)\Big\|^2 \Big]  \nonumber \\
	 \le 
	\Big(\frac{2}{\alpha} +\beta \Big) \frac{\Delta_F}{T+1} + \Big( \frac{\beta}{\eta}+2\Big) \sigma^2 \! + \frac{(\beta+2)L_{\bld{\bar \theta}}^2 \epsilon}{\lambda_0} 
\end{align} 
where we used $\mathbb{E} [\|\nabla f (\bar{\bm \theta}^t) - \bm g^\ast(\bar{\bm \theta}^t)\|_2^2] \le \sigma^2$, which holds according to Assumption \ref{as:grdestimate}. By \cite [Theorem 10]{rockafellar2009variational} and \cite{xu2019non}, we know that 
\begin{equation}
-\bm{g}^\epsilon(\bar{\bm{\theta}}^t) - \frac{1}{\alpha}(\bar{\bm{\theta}}^{t+1}-\bar{\bm{\theta}}^t) \in \partial \bar r(\bar{\bm{\theta}}^{t+1}) 
\end{equation}
which gives
\begin{align}
\nabla f(\bar{\bm{\theta}}^{t+1}) - \bm{g}^\epsilon  & (\bar{\bm{\theta}}^t) - \frac{1}{\alpha}(\bar{\bm{\theta}}^{t+1}-\bar{\bm{\theta}}^t) \in  \nonumber \\ & \nabla f(\bar{\bm{\theta}}^{t+1}) + \partial \bar r(\bar{\bm{\theta}}^{t+1})  = \partial F(\bar{\bm{\theta}}^{t+1})
\end{align}
replace this in the left hand side of \eqref{eq:bnd_sbdif}, 
and recalling the definition of distance, it holds that  
\begin{align*}
	\mathbb{E}\big[\textrm{dist}(\bm{0},  \partial \hat{F}(\bar{\bld \theta}^{t'}))\big]&\le\!   \big(\frac{2}{\alpha}\! +\!\beta \big) \frac{\Delta_F}{T}\! +\! \big( \frac{\beta}{\eta}\!+\!2\big) \sigma^2 \!+\! \frac{(\beta+2)L_{\bld{\bar \theta z}}^2 \epsilon}{\lambda_0}
\end{align*}
where $t'$ is randomly drawn from $ t'\in\{1,2,\ldots, T+1\}$, which concludes the proof.

\subsection{Proof of Theorem \ref{alg:spgda}}
\label{thm:spgda}
Instead of resorting to an oracle to obtain $\epsilon$-optimal solver for surrogate loss, here we utilize a single step stochastic gradient ascent with mini-batch size $M$ to solve the maximization step. Consequently, the successive updates become 
\begin{equation}
\bar{\bm{\theta}}^{t+1} = \textrm{prox}_{\alpha_t r}
\big(\bar{\bm{\theta}}^{t} -  \alpha_t \bm g^{t}(\bar{\bm \theta}^t) \big)  
\end{equation}
where $\bm g^{t}(\bar{\bm \theta}^t)  := \frac{1}{M} \sum_{m=1}^{M} \bm g (\bar{\bm \theta}^{t}, \bm \zeta_m^{t}; \bm z_m)$. Let us define $\bm \delta(\bar{\bm{\theta}}^t) = \nabla f(\bar{\bm{\theta}}^t) - \bm g^t(\bar{\bm{\theta}}^t)$, and use $L_f$-smoothness of $f(\bar{\bm{\theta}})$, to obtain  
\begin{align} 
f(\bar{\bm \theta}^{t+1}) & \le f(\bar{\bm{\theta}}^{t}) + \big \langle \nabla f(\bar{\bm{\theta}}^t), \bar{\bm{\theta}}^{t+1} - \bar{\bm{\theta}}^{t} \big \rangle \nonumber \\ 
& \quad + \frac{L_f}{2} \| \bar{\bm{\theta}}^{t+1} - \bar{\bm{\theta}}^{t} \|^2 \nonumber \\
& \le f(\bar{\bm{\theta}}^{t}) + \big\langle \bm  g^t (\bar{\bm{\theta}}^t)+ \bm  \delta(\bar{\bm{\theta}}^t) , \bar{\bm{\theta}}^{t+1} - \bar{\bm{\theta}}^{t}\big \rangle \nonumber \\ 
& \quad + \frac{L_f}{2} \| \bar{\bm{\theta}}^{t+1} - \bar{\bm{\theta}}^{t} \|^2. 
\label{eq:fsmooth}
\end{align}
Now, substitute $ \bar{\bm{\theta}}^{t+1} \rightarrow \bm  u $, $  \bm{\theta}^t \rightarrow \bm{y}$, and $\bar{\bm{\theta}}^t - \alpha_t \bm  g^t (\bar{\bm{\theta}}^t) \rightarrow \bm  x$ in \eqref{eq:prox_prop}, to arrive at 
\begin{equation*}
\big \langle \bar{\bm{\theta}}^t - \alpha_t \bm  g^t (\bar{\bm{\theta}}^t) - \bar{\bm{\theta}}^{t+1} , \bar{\bm{\theta}}^t - \bar{\bm{\theta}}^{t+1}\big \rangle \le \alpha_t \bar{r}(\bar{\bm{\theta}}^t) - \alpha_t \bar{r}(\bar{\bm{\theta}}^{t+1}) 
\end{equation*}
which leads to 
\begin{equation}
\big \langle \bm g^t (\bar{\bm{\theta}}^t), \bar{\bm{\theta}}^{t+1} - \bar{\bm{\theta}}^t \big \rangle \le \bar{r}(\bar{\bm{\theta}}^t) - \bar{r}(\bar{\bm{\theta}}^{t+1}) - \frac{1}{\alpha_t} \| \bar{\bm{\theta}}^{t+1} - \bar{\bm{\theta}}^{t}\|^2.
\end{equation}
Substituting this into \eqref{eq:fsmooth}, gives
\begin{align*}
f(\bar{\bm{\theta}}^{t+1}) & \le f(\bar{\bm{\theta}}^{t}) + \big \langle \bm  \delta(\bar{\bm{\theta}}^t), \bar{\bm{\theta}}^{t+1} - \bar{\bm{\theta}}^{t}\big \rangle + \frac{L_f}{2} \| \bar{\bm{\theta}}^{t+1} - \bar{\bm{\theta}}^{t}\|^2 \nonumber \\ 
& \quad + \bar{r}(\bar{\bm{\theta}}^t) - \bar{r}(\bar{\bm{\theta}}^{t+1}) - \frac{1}{\alpha_t} \|\bar{\bm{\theta}}^{t+1} - \bar{\bm{\theta}}^{t}\|^2.
\end{align*}
Using $F(\bm \theta) = f(\bm \theta) + \bar{r}(\bm \theta)$, we have 
\begin{align} \nonumber 
F(\bar{\bm{\theta}}^{t+1}) -  F(\bar{\bm{\theta}}^{t}) & \le \big\langle \bm  \delta(\bar{\bm{\theta}}^t), \bar{\bm{\theta}}^{t+1} - \bar{\bm{\theta}}^{t}\big \rangle \nonumber \\ & + \Big( \frac{L_f}{2}- \frac{1}{\alpha_t} \Big) \| \bar{\bm{\theta}}^{t+1} - \bar{\bm{\theta}}^{t}\|^2.
\end{align}
Using Young's inequality $ \big\langle \bm  \delta(\bar{\bm{\theta}}^t), \bar{\bm{\theta}}^{t+1} - \bar{\bm{\theta}}^{t}\big \rangle \le \frac{1}{2} \|\bm  \delta(\bar{\bm{\theta}}^t) \|^2+\frac{1}{2}\|\bar{\bm{\theta}}^{t+1} - \bar{\bm{\theta}}^{t} \|^2$ gives  
\begin{align} 
F(\bar{\bm{\theta}}^{t+1}) -  & F(\bar{\bm{\theta}}^{t}) \le \Big(\frac{L_f \!+ \!1}{2}- \frac{1}{\alpha_t} \Big) \|\bar{\bm{\theta}}^{t+1} - \bar{\bm{\theta}}^{t}\|^2 + \frac{\|\bm  \delta(\bar{\bm{\theta}}^t)\|^2}{2}.
\label{eq:bnd_itrsF}
\end{align}
Adding the term $\big \langle \bar{\bm{\theta}}^{t+1} - \bar{\bm{\theta}}^t, \nabla f(\bar{\bm{\theta}}^{t+1}) \big \rangle$ to both sides in \eqref{eq:bnd_itrsF} and simplifying terms yields   
\begin{align}
\nonumber
& \big \langle \bar{\bm{\theta}}^{t+1} - \bar{\bm{\theta}}^t,  \bm g^t (\bar{\bm{\theta}}^t) - \nabla f(\bar{\bm{\theta}}^{t+1})\big \rangle \nonumber \\ &  \le -\Big (\frac{1}{2 \alpha_t} - \frac{L_f}{2} \Big) \|\bar{\bm{\theta}}^{t+1} - \bar{\bm{\theta}}^{t}\|^2 + F(\bar{\bm{\theta}}^{t}) - F(\bar{\bm{\theta}}^{t+1})  \nonumber 
\\ & \quad \,
 - \big \langle\bar{\bm{\theta}}^{t+1} - \bar{\bm{\theta}}^t, \bm \nabla f (\bar{\bm{\theta}}^{t+1}) - \nabla f(\bar{\bm{\theta}}^t)\big \rangle.
\end{align}

Completing the squares yields
\begin{align}
\nonumber
& \big\|\bm g^t (\bar{\bm{\theta}}^t) - \nabla f(\bar{\bm{\theta}}^{t+1}) + \frac{1}{\alpha_t} (\bar{\bm{\theta}}^{t+1} - \bar{\bm{\theta}}^t)\big\|^2 \\ 
& \le  \|\bm g^t (\bar{\bm{\theta}}^t) - \nabla f(\bar{\bm{\theta}}^{t+1})\|^2 
+ \frac{1}{\alpha_t^2} \|\bar{\bm{\theta}}^{t+1} - \bar{\bm{\theta}}^t\|^2
 \nonumber \\ 
& \quad + \Big(\frac{L_f}{\alpha_t} - \frac{1}{\alpha_t^2}\Big) \|\bar{\bm{\theta}}^{t+1} - \bar{\bm{\theta}}^{t}\|^2 + \frac{2 (F(\bar{\bm{\theta}}^{t}) - F(\bar{\bm{\theta}}^{t+1}))}{\alpha_t} \nonumber \\ 
& \quad - \frac{2}{\alpha_t} \big \langle \bar{\bm{\theta}}^{t+1} - \bar{\bm{\theta}}^t, \nabla f (\bar{\bm{\theta}}^{t+1}) - \nabla f(\bar{\bm{\theta}}^t)\big \rangle \nonumber \\ 
& \le 2 \|\bm g^t (\bar{\bm{\theta}}^t) - \nabla f(\bar{\bm{\theta}}^{t})\|^2 + 2 \|\nabla f (\bar{\bm{\theta}}^t) - \nabla f(\bar{\bm{\theta}}^{t+1})\|^2  
\nonumber \\ 
& \quad + \frac{1}{\alpha_t^2} \|\bar{\bm{\theta}}^{t+1} - \bar{\bm{\theta}}^t\|^2 + \Big(\frac{L_f}{\alpha_t} - \frac{1}{\alpha_t^2}\Big) \|\bar{\bm{\theta}}^{t+1} - \bar{\bm{\theta}}^{t}\|^2 \nonumber \\ 
& \quad + \frac{2 (F(\bar{\bm{\theta}}^{t}) - F(\bar{\bm{\theta}}^{t+1}))}{\alpha_t} \nonumber \\ 
& \quad - \frac{2}{\alpha_t} \langle  \bar{\bm{\theta}}^{t+1} - \bar{\bm{\theta}}^t , \nabla f (\bar{\bm{\theta}}^{t+1}) - \nabla f(\bar{\bm{\theta}}^t) \rangle \nonumber \\ 
& \le 2 \|\bm g^t (\bar{\bm{\theta}}^t) - \nabla f(\bar{\bm{\theta}}^{t})\|^2 + 2 L_f^2 \|\bar{\bm{\theta}}^{t+1} - \bar{\bm{\theta}}^{t}\|^2   \nonumber \\ 
& \quad + \frac{1}{\alpha_t^2} \|\bar{\bm{\theta}}^{t+1} - \bar{\bm{\theta}}^{t}\|^2 + \Big(\frac{L_f}{\alpha_t} - \frac{1}{\alpha_t^2}\Big) \|\bar{\bm{\theta}}^{t+1} - \bar{\bm{\theta}}^{t}\|^2 \nonumber \\ 
&  \quad + \frac{2 (F(\bar{\bm{\theta}}^{t}) - F(\bar{\bm{\theta}}^{t+1}))}{\alpha_t} + \frac{2 L_f}{\alpha_t} \|\bar{\bm{\theta}}^{t+1} - \bar{\bm{\theta}}^{t}\|^2\nonumber \\ 
& 
\le  2 \|\bm g^t (\bar{\bm{\theta}}^t) - \nabla f(\bar{\bm{\theta}}^{t})\|^2 + \frac{2(F(\bar{\bm{\theta}}^{t}) - F(\bar{\bm{\theta}}^{t+1}))}{\alpha_t} \nonumber \\ 
& \quad + \frac{3L_f + 2L_f^2 \alpha_t }{\alpha_t} \|\bar{\bm{\theta}}^{t+1} - \bar{\bm{\theta}}^{t}\|^2.
\label{eq:subdifbound}
\end{align}

Recalling that $\bm \delta(\bar{\bm{\theta}}^t) = \nabla f(\bar{\bm{\theta}}^t) - \bm g^t(\bar{\bm{\theta}}^t)$, we bound the the first term as follows
\begin{align}\label{eq:graderrorbound}
\mathbb E \Big[\|\bm g^{t} (\bar{\bm{\theta}}^{t})  - & \nabla f(\bar{\bm{\theta}}^{t})\|^2\big|\bm\theta^t\Big] \nonumber \\ 
& =\mathbb E \Big[\left \| \bm g^{\ast}(\bar{\bm{\theta}}^{t})  - \nabla f(\bar{\bm{\theta}}^{t}) + \bm \delta^{t} \right \|^2\big|\bm\theta^t\Big]  \nonumber \\ 
& =  \| \bm g^{\ast}(\bar{\bm{\theta}}^{t})  - \nabla f(\bar{\bm{\theta}}^{t}) \|^2 + \| \bm \delta^{t}  \|^2 \nonumber \\ 
& \quad + 2 \mathbb E \Big[\left \langle \bm g^{\ast}(\bar{\bm{\theta}}^{t})  - \nabla f(\bar{\bm{\theta}}^{t}) ,\bm \delta^{t} \right \rangle\big|\bm{\theta}^t\Big]
\end{align}
where the third equality is obtained by expanding the square term and using $\mathbb E \big[\langle \bm g^{\ast}(\bar{\bm{\theta}}^{t})  - \nabla f(\bar{\bm{\theta}}^{t}) ,\bm \delta^{t} \rangle  \big| \bar{\bm{\theta}}^t \big] = \bm 0$. We further bound right hand side here as follows. Recalling that $\bm \delta^{t} =  \frac{1}{M} \sum_{m=1}^{M} \bm g (\bar {\bm \theta}^{t}, \bm \zeta_m^{t}; \bm z_m)  - \bm g^{\ast}(\bar{\bm \theta}^{t})$, where  $\bm g^\ast (\bm \theta^t) := \frac{1}{M} \sum_{m=1}^M \nabla_{\bar{\bm{\theta}}} \psi(\bar{\bm{\theta}}^{t}, \bm  \zeta_m^{\ast t}; \bm  z_m)$, it holds that  
\begin{align}
& \mathbb{E}\Big[\|\bm \delta^{t}\|^2\big|\bar{\bm{\theta}}^t, \bm \zeta_m^t\Big] \nonumber \\ 
&= \mathbb{E}\Big[\Big\| \frac{1}{M} \sum_{m=1}^{M}\Big[ \bm g (\bar{\bm \theta}^{t}, \bm \zeta_m^{t}; \bm z_m)  -\bm{g}^\ast(\bar{\bm{\theta}}^t) \big]\Big\|^2\Big|\bar{\bm{\theta}}^t, \bm \zeta_m^t \Big]\nonumber\\
&=\frac{1}{M^2}\sum_{m=1}^M \mathbb{E} \Big[\big\| \nabla_{\bar{\bm{\theta}}} \psi(\bar{\bm{\theta}}^{t}, \bm  \zeta_m^{t}; \bm  z_m) \nonumber \\ 
& \quad -\nabla_{\bar{\bm{\theta}}} \psi(\bar{\bm{\theta}}^{t}, \bm  \zeta_m^{\ast t} ; \bm  z_m)\big\|^2\big|\bar{\bm{\theta}}^t, \bm \zeta_m^t \Big]
\nonumber\\
&
\le 
\frac{L^2_{\bm \theta \bm z}}{M^2}   \sum_{m=1}^{M} \!\left\| \bm \zeta_m^{t}  - {\bm \zeta^{\ast t}_m} \right \|^2
\label{eq:gtgstar}
\end{align} 
where the second equality is because of i.i.d samples $\{\bm{z}_m\}_{m=1}^M$, and last inequality holds due to Lipschitz smoothness of $\psi(\cdot)$. Since $\bm{\zeta}_m^t$ is obtained by a single gradient ascent update over a $\mu$-strongly concave function, we have that 
\begin{equation}
\frac{L^2_{\bm \theta \bm z}}{M^2}   \sum_{m=1}^{M} \!\left\| \bm \zeta_m^{t}  - {\bm \zeta^{\ast t}_m} \right \|^2 \le \frac{L^2_{\bm \theta \bm z}}{M}  \Big[\! \left(1-\alpha_t \mu \right) D^2 + \alpha_t^2 B^2 \Big]
\end{equation}
where $D$ is the feasible set diameter and $\alpha_t>0$ is the step size. The following holds for the expected error term 
\begin{equation}
\mathbb{E}\Big[\|\bm \delta^{t}\|^2\big|\bar{\bm{\theta}}^t, \bm \zeta_m^t\Big] \le \frac{L^2_{\bm \theta \bm z}}{M}  \Big[\! \left(1-\alpha_t \mu \right) D^2 + \alpha_t^2 B^2 \Big] 
\end{equation}
using this in \eqref{eq:graderrorbound} to arrive at 
\begin{align}
\mathbb E \Big[\|\bm g^{t} (\bar{\bm{\theta}}^{t}) - & \nabla f(\bar{\bm{\theta}}^{t})\|^2\big|\bm\theta^t\Big] \le 2 \| \bm g^{\ast}(\bar{\bm{\theta}}^{t})  - \nabla f(\bar{\bm{\theta}}^{t}) \|^2 \nonumber \\ 
& \quad + \frac{L^2_{\bar{\bm{\theta}} \bm z}}{M}  \Big[\! \left(1-\alpha_t \mu \right) D^2 + \alpha_t^2 B^2 \Big]. 
\end{align}

Substituting the last inequality  into \eqref{eq:subdifbound}, to get
\begin{align}
&\quad \mathbb E \Big[ \big\|\bm g^t (\bar{\bm{\theta}}^t) - \nabla f(\bar{\bm{\theta}}^{t+1}) + \frac{1}{\alpha_t} (\bar{\bm{\theta}}^{t+1} - \bar{\bm{\theta}}^t)\big\|^2 \big| \bar{\bm{\theta}}^t \Big]\nonumber\\
& \le 4  \| \bm g^{\ast}(\bar{\bm{\theta}}^{t})  - \nabla f(\bar{\bm{\theta}}^{t}) \|^2  +\frac{3L_f + 2L_f^2 \alpha_t }{\alpha_t} \mathbb E \Big[ \|\bar{\bm{\theta}}^{t+1} - \bar{\bm{\theta}}^{t}\|^2 \big| \bar{\bm{\theta}}^t\Big]  \nonumber \\ 
& \quad+\frac{2 F(\bar{\bm{\theta}}^{t}) - 2 \mathbb E \big[ F(\bar{\bm{\theta}}^{t+1})\big| \bar{\bm{\theta}}^t  \big]}{\alpha_t} \nonumber \\ & \quad +\frac{L^2_{\bar{\bm{\theta}} \bm z}}{M}  \Big[\! \left(1-\alpha_t \mu \right) D^2 + \alpha_t^2 B^2 \Big]  .
\end{align}

Taking again expectation of both sides with respect to $\bar{\bm{\theta}}^t$
\begin{align}
&  \mathbb E  \Big\|\bm g^t (\bar{\bm{\theta}}^t) - \nabla f(\bar{\bm{\theta}}^{t+1}) + \frac{1}{\alpha_t} (\bar{\bm{\theta}}^{t+1} - \bar{\bm{\theta}}^t)\Big\|^2     \nonumber \\ 
& \le  4  \mathbb E \Big[ \| \bm g^{\ast}(\bar{\bm{\theta}}^{t})  - \nabla f(\bar{\bm{\theta}}^{t}) \|^2 \Big] + \frac{L^2_{\bar{\bm{\theta}} \bm z}}{M}  \Big[\! \left(1-\alpha_t \mu \right) D^2 + \alpha_t^2 B^2 \Big]  \nonumber \\ 
& \quad + \mathbb E \Big[ \frac{2 F(\bar{\bm{\theta}}^{t}) - 2 F(\bar{\bm{\theta}}^{t+1})}{\alpha_t} + \frac{3L_f + 2L_f^2 \alpha_t }{\alpha_t}  \|\bar{\bm{\theta}}^{t+1} - \bar{\bm{\theta}}^{t}\|^2 \Big].\label{eq:subdiffsgda}
\end{align}
Using our assumption $\mathbb E [\|\bm \psi^\ast (\bar{\bm{\theta}}^{t}, \bm \zeta_m^{t}; \bm z_m)-\nabla f(\bar{\bm{\theta}}^t)\|^2] \le \sigma^2$ and the fact that $\bm g^\ast(\bar{\bm{\theta}}^t) = \frac{1}{M} \sum_{m=1}^{M} \bm \psi (\bar{\bm{\theta}}^{t}, \bm \zeta_m^{\ast t}; \bm z_m)$, the first term on the right hand side can be bounded by $\frac{4\sigma^2}{M}$. For a fixed learning rate $\alpha>0$, summing inequalities \eqref{eq:subdiffsgda} from $t=0,\ldots,T$ yields
\begin{align}
& \frac{1}{T+1}\mathbb{E}\Big[\sum_{t=0}^T\big\|\bm g^t (\bar{\bm{\theta}}^t) - \nabla f(\bar{\bm{\theta}}^{t+1}) + \frac{1}{\alpha_t} (\bar{\bm{\theta}}^{t+1} - \bar{\bm{\theta}}^t)\big\|^2
\Big] \nonumber \\ 
& \le \frac{2}{\alpha(T+1)}\big(F(\bm{\theta^0})-\mathbb E [F(\bm\theta^T)]\big) \nonumber\\
& \quad
+\frac{3L_f + 2L_f^2 \alpha }{\alpha}  \frac{1}{T+1} \mathbb E \Big[ \sum_{t=0}^T    \|\bar{\bm{\theta}}^{t+1} - \bar{\bm{\theta}}^{t}\|^2 \Big] +
\frac{4\sigma^2}{M} \nonumber \\ 
& \quad + \frac{2L^2_{\bar{\bm{\theta}} \bm z}}{M}  \Big[(1-\alpha \mu ) D^2 + \alpha^2 B^2 \Big]  
\nonumber \\ 
& \le \frac{1}{T+1} \bigg\{\frac{2}{\alpha} + \frac{6L_f + 4L_f^2 \alpha }{[2- \alpha(L_f+\beta)]}\bigg\} ( F(\bar{\bm{\theta}}^{0}) -  \mathbb E [F(\bar{\bm{\theta}}^{T})] ) \nonumber \\ 
& \quad +\frac{2  L^2_{\bar{\bm{\theta}} \bm z}}{M}\bigg\{ 1+ \frac{ 3L_f + 2L_f^2 \alpha }{2(2- \alpha(L_f+\beta))}  \bigg\} \Big[\! \left(1-\alpha \mu \right) D^2 + \alpha^2 B^2 \Big] \nonumber \\ 
& \quad + \frac{4 \sigma^2}{M}. 
\end{align}
Replace $F(\bar{\bm{\theta}}^0) -  F(\bar{\bm{\theta}}^T)$ with $\Delta_F = F(\bar{\bm{\theta}}^0) -\inf_{\bar{\bm{\theta}}} \bm F(\bar{\bm{\theta}})$, and note that $\bm g^t (\bar{\bm{\theta}}^t) - \nabla f(\bar{\bm{\theta}}^{t+1}) + \frac{1}{\alpha_t} (\bar{\bm{\theta}}^{t+1} - \bar{\bm{\theta}}^t) \in \partial {F}(\bar{\bm{\theta}}^{t+1})$, where $\partial {F}$ denotes the set of sub gradients of $F$. It becomes clear that
\begin{align}
& \mathbb E \big[ \textrm{dist}(0, \partial F)^2 \big] \nonumber \\ 
&\le \frac{1}{T+1}\mathbb{E}\Big[\sum_{t=0}^T\Big\|\bm g^t (\bar{\bm{\theta}}^t) - \nabla f(\bar{\bm{\theta}}^{t+1}) + \frac{1}{\alpha_t} (\bar{\bm{\theta}}^{t+1} - \bar{\bm{\theta}}^t)\Big\|^2
\Big] \nonumber \\
& \le \frac{\zeta}{T+1} \Delta_F +\frac{2  L^2_{\bar{\bm{\theta}} \bm z} \nu}{N} \Big[\! \left(1-\alpha \mu \right) D^2 + \alpha^2 B^2 \Big] + \frac{4 \sigma^2}{M}
\end{align}
where $\zeta = \frac{2}{\alpha} + \frac{6L_f + 4L_f^2 \alpha }{(2- \alpha(L_f+\beta))}$ and $\nu = 1+ \frac{3L_f + 2L_f^2 \alpha }{2(2- \alpha(L_f+\beta))}$, which concludes the proof.

\bibliographystyle{IEEEtranS}
\bibliography{refs}

\end{document}